\documentclass[%
 reprint,
superscriptaddress,
 amsmath,amssymb,
 aps,
]{revtex4-2}

\usepackage{graphicx}
\usepackage{natbib}
\usepackage[resetlabels,labeled]{multibib}
\newcites{New}{Supplementary - References}
\usepackage{subfig}
\usepackage{hyperref}
\usepackage{comment}
\usepackage{amsmath, amsthm, amsfonts, amssymb} 
\usepackage{mathrsfs}  
\usepackage[english]{babel}
\makeatletter
\def\bbl@set@language#1{%
  \edef\languagename{%
    \ifnum\escapechar=\expandafter`\string#1\@empty
    \else\string#1\@empty\fi}%
  \@ifundefined{babel@language@alias@\languagename}{}{%
    \edef\languagename{\@nameuse{babel@language@alias@\languagename}}%
  }%
  \select@language{\languagename}%
  \expandafter\ifx\csname date\languagename\endcsname\relax\else
    \if@filesw
      \protected@write\@auxout{}{\string\select@language{\languagename}}%
      \bbl@for\bbl@tempa\BabelContentsFiles{%
        \addtocontents{\bbl@tempa}{\xstring\select@language{\languagename}}}%
      \bbl@usehooks{write}{}%
    \fi
  \fi}
\newcommand{\DeclareLanguageAlias}[2]{%
  \global\@namedef{babel@language@alias@#1}{#2}%
}
\makeatother

\DeclareLanguageAlias{en}{english}
\usepackage{multirow}
\usepackage{blkarray}
\usepackage{dcolumn}
\usepackage{bm}

\begin{document}

\preprint{APS/123-QED}

\title{LEARNING FUTURE TERRORIST TARGETS THROUGH TEMPORAL META-GRAPHS}
\thanks{Article published in Nature Scientific Reports. Please cite: \href{https://www.nature.com/articles/s41598-021-87709-7}{https://www.nature.com/articles/s41598-021-87709-7}}%

\author{Gian Maria Campedelli}
 \email{Corresponding author: gianmaria.campedelli@unitn.it}
\affiliation{%
Department of Sociology and Social Research ---
University of Trento, Italy
}%


\author{Mihovil Bartulovic}
%
\author{Kathleen M. Carley}
\affiliation{%
School of Computer Science --- 
Carnegie Mellon University, Pittsburgh, PA, USA
}%


\date{\today}

\begin{abstract}
In the last twenty years, terrorism has led to hundreds of thousands of deaths and massive economic, political, and humanitarian crises in several regions of the world. Using real-world data on attacks occurred in Afghanistan and Iraq from 2001 to 2018, we propose the use of temporal meta-graphs and deep learning to forecast future terrorist targets. Focusing on three event dimensions, i.e., employed weapons, deployed tactics and chosen targets, meta-graphs map the connections among temporally close attacks, capturing their operational similarities and dependencies.  From these temporal meta-graphs, we derive two-day-based time series that measure the centrality of each feature within each dimension over time. Formulating the problem in the context of the strategic behavior of terrorist actors, these multivariate temporal sequences are then utilized to learn what target types are at the highest risk of being chosen. The paper makes two contributions. First, it demonstrates that engineering the feature space via temporal meta-graphs produces richer knowledge than shallow time-series that only rely on frequency of feature occurrences. Second, the performed experiments reveal that Bi-directional LSTM networks achieve superior forecasting performance compared to other algorithms, calling for future research aiming at fully discovering the potential of artificial intelligence to counter terrorist dynamics.
\end{abstract}

\maketitle


\section{Introduction}

After peaking in 2014, terrorism activity worldwide has been on the decline in the last five years, as a consequence of several major defeats suffered by the Islamic State and Boko Haram,  two of the world's most prominent jihadist organizations. Nonetheless, terrorism remains a persistent threat to the population of many areas of the world. Despite the decline in attacks and fatalities, in 2018 alone terrorist attacks worldwide led to 15,952 deaths \cite{InstituteforEconomicsandPeaceGlobalTerrorismIndex2019}.

The multifaceted nature of terrorism, characterized by a myriad of ideologies, motives, actors, and objectives, poses a challenge to governments, institutions, and policy-makers around the world. Terrorism, in fact, undermines states' stability, peace, and cooperation between countries, in addition to economic development and basic human rights. Given its salience and relevance, the United Nations includes the prevention of terrorism (along with violence and crime) as a target of the sixteenth Sustainable Development Goal, which specifically frames the promotion of peaceful and inclusive societies.

Scholars have called for the development of a dedicated scientific field focusing on the computational study of conflicts, civil wars, and terrorism \cite{GuoRetoolAIforecast2018,McKendrickArtificialIntelligencePrediction2019}. However, to date, attempts to exploit artificial intelligence for such purposes have been few and scattered. Whilst terrorism remains characterized by high levels of uncertainty and unpredictability \cite{SchiermeierAttemptspredictterrorist2015}, trans-disciplinary research can help in providing data-driven solutions aimed at countering this phenomenon, exploiting the promising juncture of richer data, powerful computational models, and solid theories of terrorist behavior.

In light of this, the present study aims at bridging artificial intelligence and terrorism research by proposing a new computational framework based on meta-graphs, time-series, and forecasting algorithms. Retrieving event data from the Global Terrorism Database, we focus on all the attacks that occurred in Afghanistan and Iraq from 2001 to 2018 and construct two-day-based meta-graphs representing the operational connections emerging from three event dimensions: utilized weapons, deployed tactics, and chosen targets. Once meta-graphs are created, we derive time-series mapping the centrality of each feature in each dimension. The generated time series are then utilized to learn the existing recurring patterns between operational features to forecast the next most likely central - and therefore popular - targets. 

A baseline approach assuming no changes in terrorist dynamics over time and five deep learning models (i.e., Feed-Forward Neural Networks, Long Short-Term Memory Networks, Convolutional Neural Networks, Bidirectional Long Short-Term Memory Networks, and Convolutional Long Short-Term Memory Networks) are assessed in terms of forecasting performance. The outcomes are compared in relation to Mean Squared Error and two metrics that we introduce for this case study: Element-wise and Set-wise Accuracy. Furthermore, our graph-based feature engineering framework is compared against models that exploit shallow time-series simply reporting the aggregate count of each tactic, weapon, and target in each two day-based time unit. The comparison aims at demonstrating that incorporating operational inter-dependencies through network metrics provides more information than merely considering event characteristics as independent from one another.

The statistical results signal that time-series gathered from temporal meta-graphs are better suited than shallow time-series for forecasting the next most central targets. Furthermore,  Bidirectional Long Short-Term Memory networks achieve higher results compared to other modeling alternatives in both datasets. Forecasting outcomes is promising and stimulates future research designed to exploit the strength of computational sciences and artificial intelligence to study terrorist events and behaviors. Our work and presented outcomes pave the path for further collaboration among different disciplines to combine the practical necessity to forecast and predict as well as the need to theoretically and etiologically understand how terrorist groups act to strategically maximize their payoffs.

\section{Background}
\subsection{Related Work}
The study of terrorist targets holds a prominent role in the literature: beyond theoretical relevance, shining a light on individuals or entities at high risk of being hit can indeed help in designing prevention policies and allocating resources to protect such targets \cite{ClarkeOutsmartingTerrorists2006,BierChoosingWhatProtect2007}. 

Studies investigating the characteristics and dynamics behind terrorist target selection have mainly employed traditional time-series methods, relying on yearly- or monthly-based observations \cite{EndersTransnationalTerrorismBecoming2000a,EndersPatternsTransnationalTerrorism2002a,AsalSoftestTargetsStudy2009b,SantifortTerroristattacktarget2013a},  mostly framing research in the spirit of inference, rather than forecasting or prediction.  These works highlighted the high-level patterns occurring globally, signaling how, over the decades, terrorist actors have substantially changed their operational and strategical behaviors. Nonetheless, these analytical approaches have limited ability to provide actionable knowledge for practically solving the counter-terrorism problem of resource allocation and attack prevention, given their meso or macro temporal focus.   

More recently, an increasing use of computational approaches favored by a higher availability of data fostered the diffusion of  works that focused on temporal micro scales. Scholars have applied computational models investigating attack sequences, analyzing the spatio-temporal concentration of terrorist events and testing novel algorithmic solutions aimed at predicting future activity, with a particular emphasis on hotspots or violent eruptions.  Among the tested algorithmic solutions are the use of point process modeling \cite{Zammit-MangionPointprocessmodelling2012a, TenchSpatiotemporalpatternsIED2016f,  ClarkModelingestimationselfexciting2018a}, network-based approaches \cite{DesmaraisForecastinglocationaldynamics2013b, CampedelliComplexNetworksTerrorist2018}, Hidden Markov models \cite{PetroffUsingHiddenMarkov2013}, near-repeat analysis \cite{ChuangLocalalliancesrivalries2019}, and early-warning statistical solutions using partial attack sequences \cite{YangQuantifyingfuturelethality2019}.  In this computationally-intensive strand of research, the attention on terrorist targets, however, has been overlooked. 

Overall, most literature has modeled terrorist attacks treating all events without discriminating them by their substantial features. Yet, this simplification greatly underestimates the multi-layered complexity of terrorist dynamics. Besides being patterned in their temporal characterization, terrorist attacks may follow patterns also in their essential operational nature \cite{Campedellicomplexnetworksapproach2019a, Poloqualityterroristviolence2020}. Ignoring this information and assuming all attacks are uni-dimensional fail to consider the hidden connections between temporally close events and the recurring operational similarities of distinct campaigns or strategies. 

Machine and deep learning algorithms can help to overcome the limitations of the extant research in terms of paucity of attention to the fine-grained temporal analysis of terrorist targets, answering the call for scientific initiatives that should develop stronger connections between methodology and theory, rather than merely privileging one of the two \cite{BakkerForecastingTerrorismNeed2012}. The power, flexibility, ability to detect and handle non-linearity of these algorithmic architectures represent promising advantages that the field of terrorism research should explore.  In the last years, few studies have attempted to exploit the strengths of artificial intelligence in this domain. Among these, Liu and colleagues \cite{LiuPredictingNextLocation2016} presented a novel recurrent model with spatial and temporal components, and used data on terrorist attacks as one of two distinct experiments to evaluated the method's performance. However, the authors do not address how data have been processed before the proper modeling part, nor sufficiently clarify the implications of their forecasts, largely affecting the theoretical value of the experiment for terrorism research purposes. Ding and co-authors \cite{DingUnderstandingdynamicsterrorism2017}, instead, used data on terrorist attacks from 1970 to 2015 to forecast event locations in 2016, comparing the ability of three different machine learning classifiers in solving the task. While the authors interestingly combine several data sources trying to connect the methodological aspect with theory, the yearly scale of their predictions limits the usefulness of the results from a practical point of view, in line with issues already described in the literature employing more traditional statistical approaches. 

More recently, Jain et al. \cite{JainToyModelStudy2019} presented the results of a study aimed at highlighting the promises of Convolutional Neural Networks in predicting long-term terrorism activity. However, the authors do not employ existing real-world data, but instead evaluate their approach using artificially generated data. Furthermore, the models only employ univariate signals, excluding potential correlated signals that may impact forecasts. 

In light of the sparsity and scarcity of works in this domain,  this work on the one hand proposes a computational framework aimed at bridging the two disciplines of terrorism research and artificial intelligence for forecasting most likely targets and fostering the use of machine and deep learning for social good, using real world data at a fine-grained temporal resolution. On the other hand, we seek to contribute to the theoretical study of terrorism by framing the problem in the context of the strategic theories of terrorist behavior.

\subsection{Theoretical Framework}
Different theories have been proposed to describe and explain terrorist actions. These can be mainly divided into three perspectives: (1) psychological, (2) organizational and (3) strategical. Psychological theories of terrorism aim at explaining the individual causes leading to join terrorist actions and are mostly concerned with considerations covering motivations, individual drivers, and personal traits. Organizational theories, in turn, focus on the internal structure and the formal symbolism of each group as a way to read their behavior. Finally, strategic theories --- which are derived from rationalist philosophy --- address terrorist groups' decision-making and originate in the study of conflicts. Within this latter field, Schelling \cite{SchellingStrategyConflict1980} posited that the parties engaging in a conflict are adaptive strategic agents:  they hence try to find the most suitable ways to win, ruling out the opponent, as in a game or a contest. This straightforward consideration
has been widely adopted by terrorism researchers who have formalized terrorism
as an instrumental type of activity carried out to achieve a given set of long and
short-run objectives \cite{CorsiTerrorismDesperateGame1981}.
As noted by McCormick \cite{McCormickTerroristDecisionMaking2003}, terrorist groups are organizations that aim at maximizing their expected political returns or minimizing the expected costs related to a set of objectives. Notably, besides this adaptive and adversarial characteristic, the strategic frame assumes that terrorist groups act with a collective rationality \cite{CrenshawTheoriesterrorismInstrumental1987, Sandlercalculusdissentanalysis1988}: a terrorist group can be thought of as a unique actor, existing a unitary entity \textit{per se}, in spite of its distinct internal components. Although this assumption simplifies reality, as terrorist groups can be structured in very different ways and these organizational features may impact decision-making processes, when considering historical events and their multidimensional characteristics, the assumption of collective rationality originated from Schelling in his studies on conflict adaptivity holds and actually helps in interpreting the life-cycle and behavior of a group. 

Many constraints severely limit the strategic decision-making of a group (i.e., limited manpower) \cite{DrakeTerroriststargetselection1998}, and such constraints have an impact on the type of attacks (as the ultimate and visible step of a decision making process) that a terrorist group will plot. The strategic theoretical approach helps in unfolding some of the visible dynamics that data can reveal, including behavioral variations in combinations of tactics, weapons, and targets \cite{McCormickTerroristDecisionMaking2003}.

Recently, empirical research has corroborated the strategical perspective, showcasing for instance that terrorist violence follows specific patterns. To exemplify, terrorism is often characterized by self-excitability and self-propagation \cite{PorterSelfexcitinghurdlemodels2012,TenchSpatiotemporalpatternsIED2016f,ClarkModelingestimationselfexciting2018a}. The occurrence of an attack increases the probability of subsequent attacks in the same area within a limited time window, similar to what happens with earthquakes and their aftershocks, as a way to rationally maximize the inflicted damages of the attack waves. Nonetheless, empirical evaluation of whether the non-random nature of attacks can be extended also to events' operational characteristics is lacking.

The intuition behind this work builds on these theoretical prepositions and seek to further scrutinize their ability to shed light on terrorism: we hypothesize that, given the complex adaptive and strategic decision-making processes in terrorist violence, we can exploit the temporal multi-dimensionality of the hidden operational connections among temporally close events to learn what type of targets will be at highest risk of being hit in the immediate future. 

\section{Data}
The analyses in this work rely on data drawn from the Global Terrorism Database (GTD), maintained by the START research center at the University of Maryland \cite{LaFreeIntroducingGlobalTerrorism2007}. The GTD is the world's most comprehensive and detailed open-access dataset on terrorist events and START releases an updated version of the dataset every year. The dataset includes now data on more than  200,000 real-world events. To be included in the dataset, an event has to meet specific criteria \cite{STARTGTDCodebookInclusion2017}. These criteria are divided into two different levels. 

There are primarily three first-level criteria that all have to be verified. These are related to the (1) intentionality and the violence (or immediate threat of violence) of the incident and (2) the sub-national nature of terrorist actors. There are also three second-level criteria, but the condition is that at least two of them are respected. Second level criteria relate to (1) the specific political, economic, religious, or social goal of each act, (2) the evidence of an intention to coerce, intimidate or convey messages to larger audiences than the immediate victims, (3) the context of action which has to be outside of legitimate warfare activities. Finally, although an event respects these two levels and is included in the dataset, an additional filtering mechanism (variable \textit{doubter}) is introduced to control for conflicting information or acts that may not be of exclusive terrorist nature. Each event is associated with dozens of variables, mapping geographic and temporal information, event characteristics, consequences in terms of fatalities and economic damages, and attack perpetrators. 

Besides considering the information on the country where an attack has occurred and the day in which the attack was plotted, this work specifically considers three core dimensions describing each event, namely (1) tactics, (2) weapons, and (3) targets. Tactics, weapons and targets that have been rarely chosen or employed in terrorist attacks occurred in Afghanistan and Iraq have been excluded by the analysis. For a detailed explanation of the rationale of this filtering, see the Supplementary Material. Descriptive statistics on terrorist attacks in both countries are reported in Table \ref{desc}.

\paragraph{Tactics}
In the GTD, each attack can be characterized by up to three different tactics. Specifically, a single attack may be plotted using a mix of different tactics, and this information generally pinpoints a certain amount of logistical complexity. In the period under consideration, attacks in Afghanistan and Iraq have deployed using ``Bombing/Explosion'', ``Hijacking'', ``Armed Assault'', ``Facility/Infrastructure Attack'', ``Assassination'', ``Hostage Taking (Kidnapping)'', ``Hostage Taking (Barricade Incident)'', and ``Unknown''.
\paragraph{Weapons}
For every event, the GTD records up to four different weapons. The higher the number of weapons utilized in a single attack, the higher the probability that the actor possesses a high amount of resources. In the Iraq and Afghanistan datasets, the represented weapon types are ``Firearms'', ``Incendiary'', ``Explosives'', ``Melee'', and ``Unknown''.
\paragraph{Targets}
Finally, each event can be associated to up to three different target categories. In the two datasets, the represented target types are the following: ``Private Citizens and Property'', ``Government (Diplomatic)'', ``Business'', ``Police', ``Government (General)'', ``NGO'', ``Journalists and Media'', ``Violent Political Party'', ``Religious Figures/Institutions'', ``Transportation'', ``Unknown'', ``Terrorists/Non-State Militia'', ``Utilities'', ``Military'', ``Telecommunication'', ``Educational Institution'', ``Tourists'', ``Other'', ``Food or Water Supply'', ``Airports \& Aircraft''.

\begin{table*}[!hbt]
\setlength{\tabcolsep}{4pt}
\footnotesize
\centering
\begin{tabular}{lccccccccc}
\hline
\textbf{\begin{tabular}[c]{@{}l@{}}Country/\\ Dataset\end{tabular}} & \textbf{\begin{tabular}[c]{@{}c@{}}Original N\\ of Attacks\end{tabular}} & \textbf{\begin{tabular}[c]{@{}c@{}}Filtered N\\ of Attacks\end{tabular}} & \textbf{\begin{tabular}[c]{@{}c@{}}Daily \\ Average\end{tabular}} & \textbf{\begin{tabular}[c]{@{}c@{}}Daily \\ St. Dev.\end{tabular}} & \textbf{\begin{tabular}[c]{@{}c@{}}Daily\\ Min.\end{tabular}} & \textbf{\begin{tabular}[c]{@{}c@{}}Daily\\ Max.\end{tabular}} & \textbf{\begin{tabular}[c]{@{}c@{}}N of \\ Targets\end{tabular}} & \textbf{\begin{tabular}[c]{@{}c@{}}N of \\ Weapons\end{tabular}} & \textbf{\begin{tabular}[c]{@{}c@{}}N of \\ Tactics\end{tabular}} \\ \hline
Afghanistan & 14,371 & 12,106 & 1.841 & 2.722 & 0 & 68 & 18 & 5 & 9 \\
Iraq & 25,886 & 22,764 & 3.462 & 4.670 & 0 & 107 & 20 & 5 & 8 \\ \hline
\end{tabular}
\caption{Descriptive Statistics of the Afghanistan and Iraq datasets reporting the total number of attacks occurred between 2001 and 2018 and the total number of targets, weapons and tactics represented in the considered time frame}
\label{desc}
\end{table*}

\begin{figure*}
    \centering
    \includegraphics[scale=0.3]{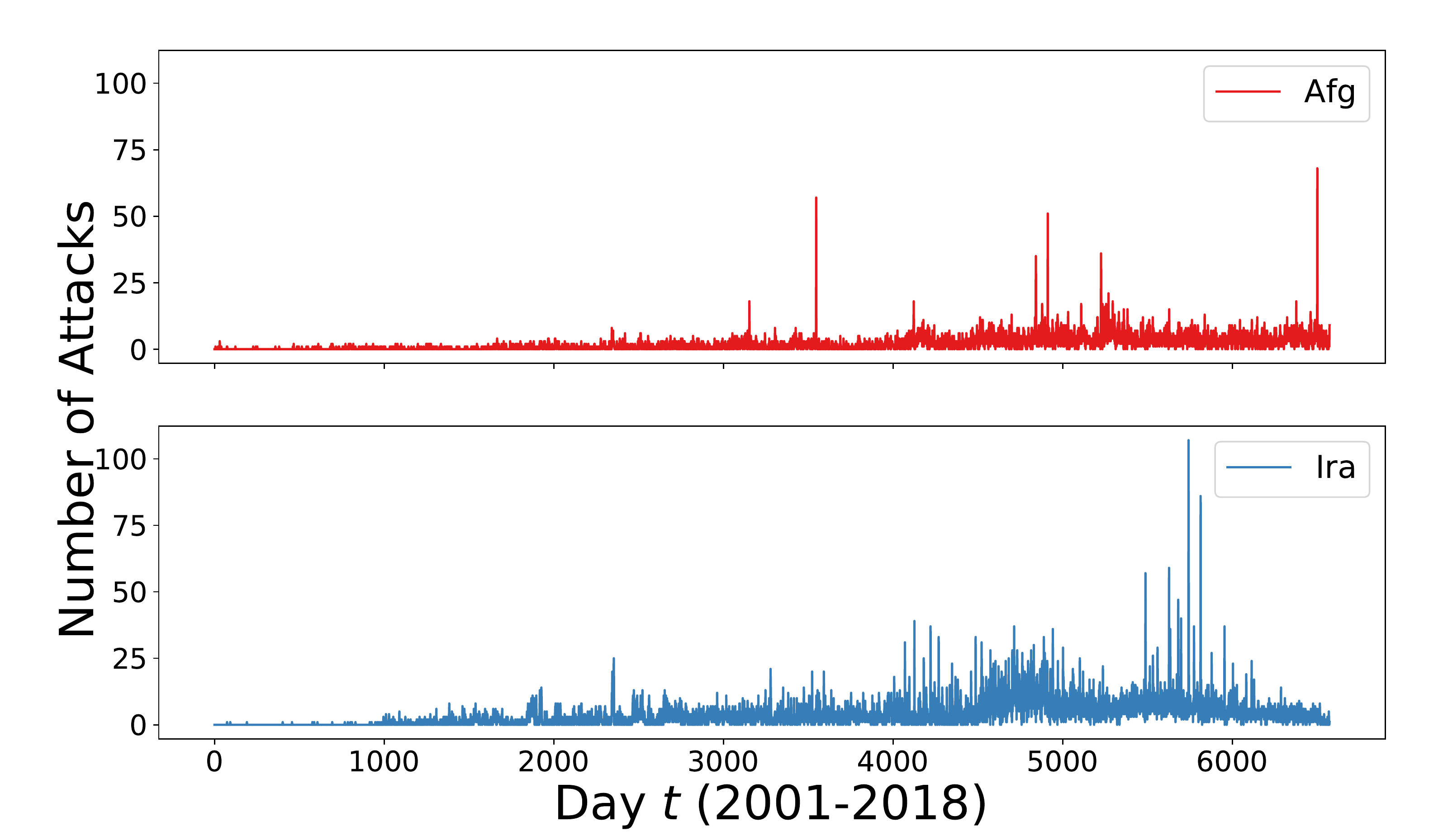}
    \caption{Time-series of terrorist attacks at the day level in Afghanistan (top) and Iraq (bottom).}
    \label{fig:daily}
\end{figure*}

The analytical experiments are performed using all the data regarding terrorist attacks that occurred in Afghanistan and Iraq from January 1st 2001 to December 31st 2018. The time series reporting the daily number of attacks in the countries under consideration are visualized in Figure \ref{fig:daily}.

\section{Temporal Meta-Graphs and Graph-derived Time-Series}
The main technical contribution of this work regards the representation of terrorist events through temporal meta graphs. The literature has shown that terrorist attacks do not occur at random. The associated core intuition is that, besides temporal clustering, there exist operational recurring patterns that can be learned to infer future terrorist actions. To capture the interconnections between events and their characteristics, framing the problem as a traditional time-series one is not sufficient. 

In light of this, we introduce a new framework that exploits the advantages of graph-derived time series. First, per each time unit weighted graphs representing the meta-connections existing within the three data dimensions under consideration (tactics, weapons, and targets) are generated. Once this step is completed, we calculate, for each dimension  and for each time unit, the normalized degree centrality of all the features. Normalized degree centrality maps the popularity of a certain weapon, tactic, or target in a given two-day temporal window, by encapsulating it in a 1-dimensional space of complex information that emerged from a clustered series of attacks. By employing graph-derived time series we couple two layers of interdependence among events: the temporal and the operational one. Centrality not only portraits a certain target popularity: it may also aid in understanding the topological structure of a given set of attacks from the operational point of view,  facilitating wide and distributed public safety strategies.

We start our data processing procedure by introducing a dataset $\mathcal{D}_{A \times z}$ that contains $|A|$ terrorist attacks and $|z|$ variables associated with each attack, exactly corresponding to the original format of the GTD. At this point, we filter out separately all the attacks that occurred in Afghanistan and Iraq in the time frame under consideration and we obtain two separate datasets: $\mathcal{D}_{t \times z }^{\mathrm{AFG}}$ and $\mathcal{D}_{t \times z }^{\mathrm{IRA}}$. The two new datasets are composed of $|t|$ observations (in this particular case one observation represents one day) and $|z|$ features. Here, the value corresponding to each feature is simply the number of times that feature was present in attacks plotted in that time unit, i.e. a single day. By doing so, we transitioned from an event-based dataset  $\mathcal{D}_{A \times z}$ to a time-based one. At this point, $\mathcal{D}_{t \times z }^{\mathrm{AFG}}$ and $\mathcal{D}_{t \times z }^{\mathrm{IRA}}$ can be subset into $m \leq |z|$ theoretical dimensions. As anticipated, the dimensions here used are $m=3$: the set of tactics features ($X$), the set of weapon features ($W$), and the set of target features ($Y$). Using the general $\mathrm{C}$ superscript indicating the country of reference, the subsetting leads to $ \mathrm{D}^{\mathrm{C}}=\left \{ \mathcal{D}_{X},\mathcal{D}_{W} ,\mathcal{D}_{Y} \right \}$. 

\begin{figure}[!hbt]
    \centering
    \includegraphics[scale=0.5]{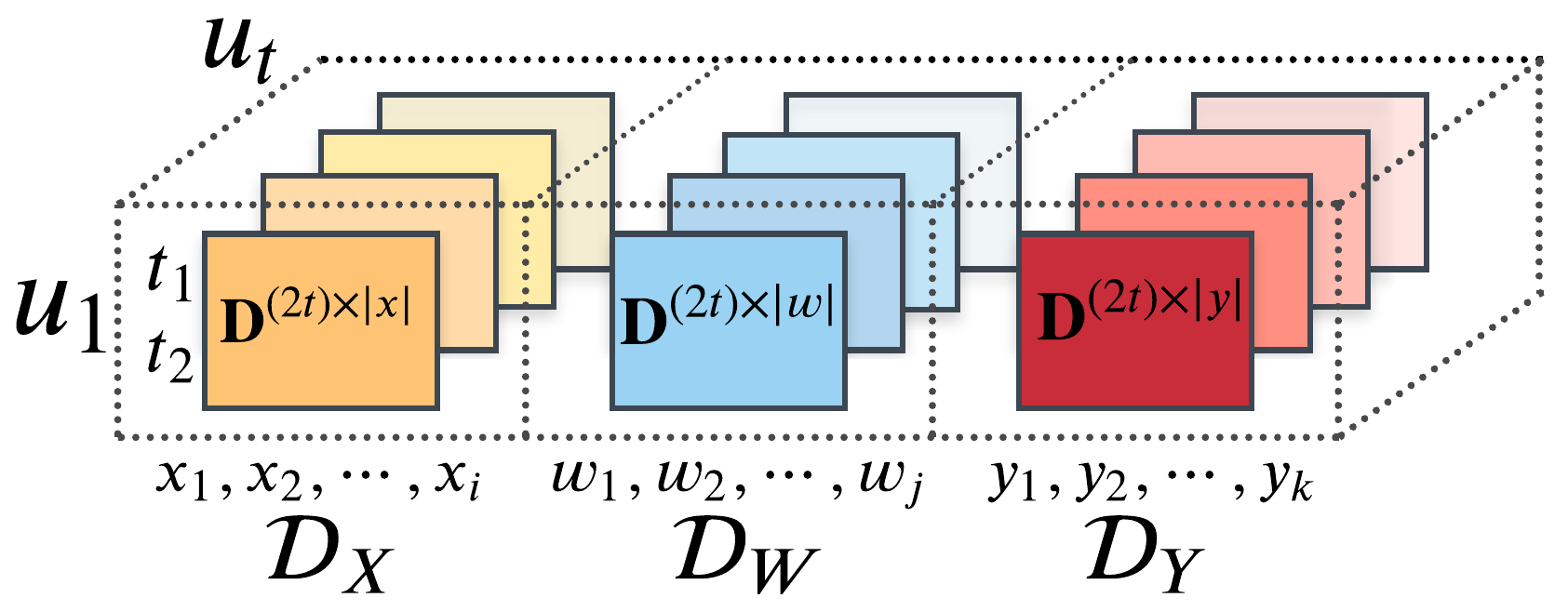}
    \caption{Graphic visualization sample of the 4d tensor resulting from the data processing.}
    \label{fig:tensor}
\end{figure}

Further, for each $\left \{ \mathcal{D}_{X},\mathcal{D}_{W} ,\mathcal{D}_{Y} \right \}$ we create $U$ temporal slices, such that $u=2t$. In other words, for each dimension, we collapse the data describing the attacks in time units made of two days each by summing the count of each feature in the same two days. The reason behind the creation of two-day-based time units is two-fold. On the one hand, relying on single day-based time series raises the risk of having overly sparse series, with very small graphs that would carry little to no relational information. On the other hand, in the real world resource allocation problems require time to be addressed, and having a forecasting system that operates day by day would produce knowledge that would be hard to transform into concrete and meaningful decisions. Thus, for this application, a two-day architecture represents a good compromise. It reduces the sparsity guaranteeing that each time input is sufficiently rich in information and it provides predictions for the next two days, such that policymakers or intelligence decision-makers  would strive less in changing resource allocation strategies too often. 
The temporal slicing leads to a 4-dimensional tensor $\mathrm{D}^{\mathrm{C}}$, as depicted in Figure \ref{fig:tensor}.
\newline

The tensor is composed by three tensors of rank 3, each representing one dimension i.e., tactics, weapons and targets. Each dimension is in turn composed by $U$ matrices. For instance, for the weapons dimension $W$, each matrix is composed by $2t$ rows and $|W|$ columns, mapping all the weapons that have at least one occurrence over the entire history. To further exemplify,  $\mathrm{\mathbf{D}_{W}[1]}$ is mapping the first temporal slice in the weapon dimension may be represented by:
\begin{equation*}
\centering
\begin{blockarray}{ccccc}
& Explosives & Firearms & \cdots & Incendiary \\
\begin{block}{c[cccc]}
  t_1 & 5 & 1 & \cdots & 0 \\
  t_2 & 1 & 0 & \cdots & 2 \\
\end{block}
\end{blockarray}    
\end{equation*}

At this point, to obtain the centrality of each feature in each matrix in the 4D-tensor, we first compute: 
\begin{equation}
    \mathbf{G}_{I}[u]=\mathbf{D}_{I}[u]^{\mathrm{T}}\mathbf{D}_{I}[u]
\end{equation}
where $I\in\{X,W,Y\}$. By multiplying the transpose $\mathbf{D}_{I}[u]^{\mathrm{T}}$ by $\mathbf{D}_{I}[u]$, we obtain a $|I|\times |I|$ square matrix whose entries represent the number of times every pair of features has been connected in the time unit under consideration. This matrix is interpreted as a meta-graph in which the connections between entities are not directly physical or tangible. Instead, the meta-graph represents a flexible abstract conceptualization aimed at linking together entities, such as particular tactics, that have been employed together in a specific time frame, within a limited set of attacks and that can be part of a logistically complex terrorist campaign.

The final step of this procedure is the computation of $\psi_{i,\mathrm{Norm}}[u]$ which is the normalized centrality of each feature $i \in I$ for each two-day temporal unit $u$. Given that $\mathbf{G}_{I}[u]$ can be interpreted as a weighted square graph,  then the weighted degree centrality of the feature $i$ in  $\mathbf{G}_{I}[u]$ is computed as:
\begin{equation}
   {\psi}_i[u] =\sum_{\substack{i \\ j\neq i}}^{|I|}\mathbf{G}_{i,j}[u]
\end{equation}
where $\mathbf{G}_{i,j}[u]$ denotes the $(i,j)$ entry of $\mathbf{G}_I[u]$. Consequently, the normalized value is obtained through:
\begin{equation}
   \psi_{i,\mathrm{Norm}}[u]  = \frac{{\psi}_i[u]}{\max_{i\in I}  {\psi}_i[u]}
\end{equation}
Normalizing the degree centrality allows to relatively compare the importance of each feature across time units that may present high variation in terrorist activity. 

\begin{figure}[!hbt]
    \centering
    \includegraphics[scale=.5]{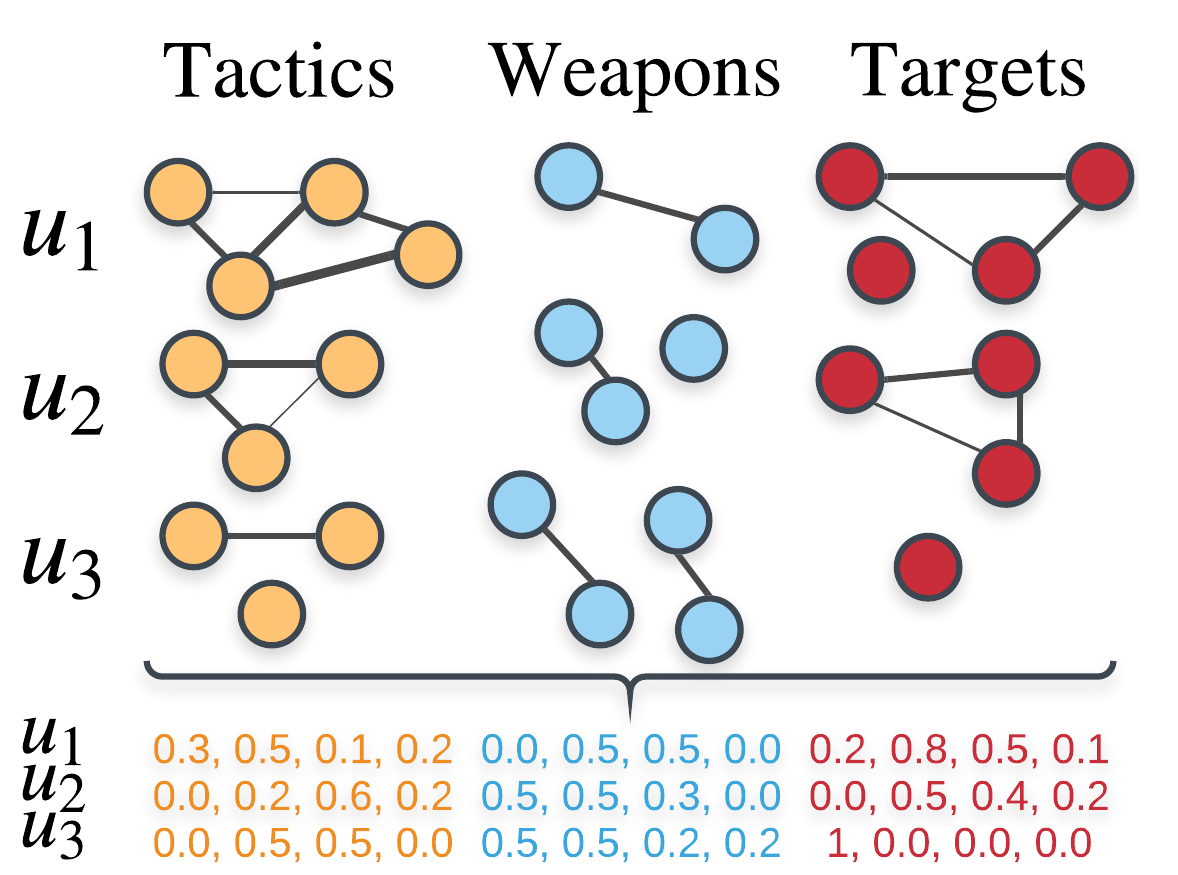}
    \caption{Sample visual depiction of the transformation of temporal meta-graphs in each dimension across $u$ time-units in multivariate time-series capturing the normalized degree centrality of each feature in each dimension across the same $u$ time units.}
    \label{fig:temporalmeta}
\end{figure}

Finally, this leads to the creation of multivariate time-series in the form:
\begin{equation}
    \Psi[U] = \left \{ \psi_{i,{\mathrm{Norm}}}[u]\right \}_{u=0}^{U}, i=1,2,...,F = |I|
\end{equation}
where $F$ is equal to the total number of features across all dimensions $|X|+|W|+|Y|$. For instance, in Afghanistan during the 2001-2018 time period attacks involved 5 weapons types, 9 tactics, and 18 targets making $F=32$. Each $\psi_{i_{\mathrm{Norm}}}[u]$ maps the relative importance of each feature in its respective dimension in a specific $u$ (a sample visualization of this process is reported in Figure \ref{fig:temporalmeta}). Instead of only using the simple counts of occurrences of each feature, the computed centrality value embeds how prevalent was a specific feature compared to the others, taking into account the meta-connections resulting from the complex logistic operations put in place by the terrorist actors active in Afghanistan and Iraq. 

\begin{figure}[!hbt]
    \centering
    \subfloat[\centering Afghanistan]{{\includegraphics[scale=.5]{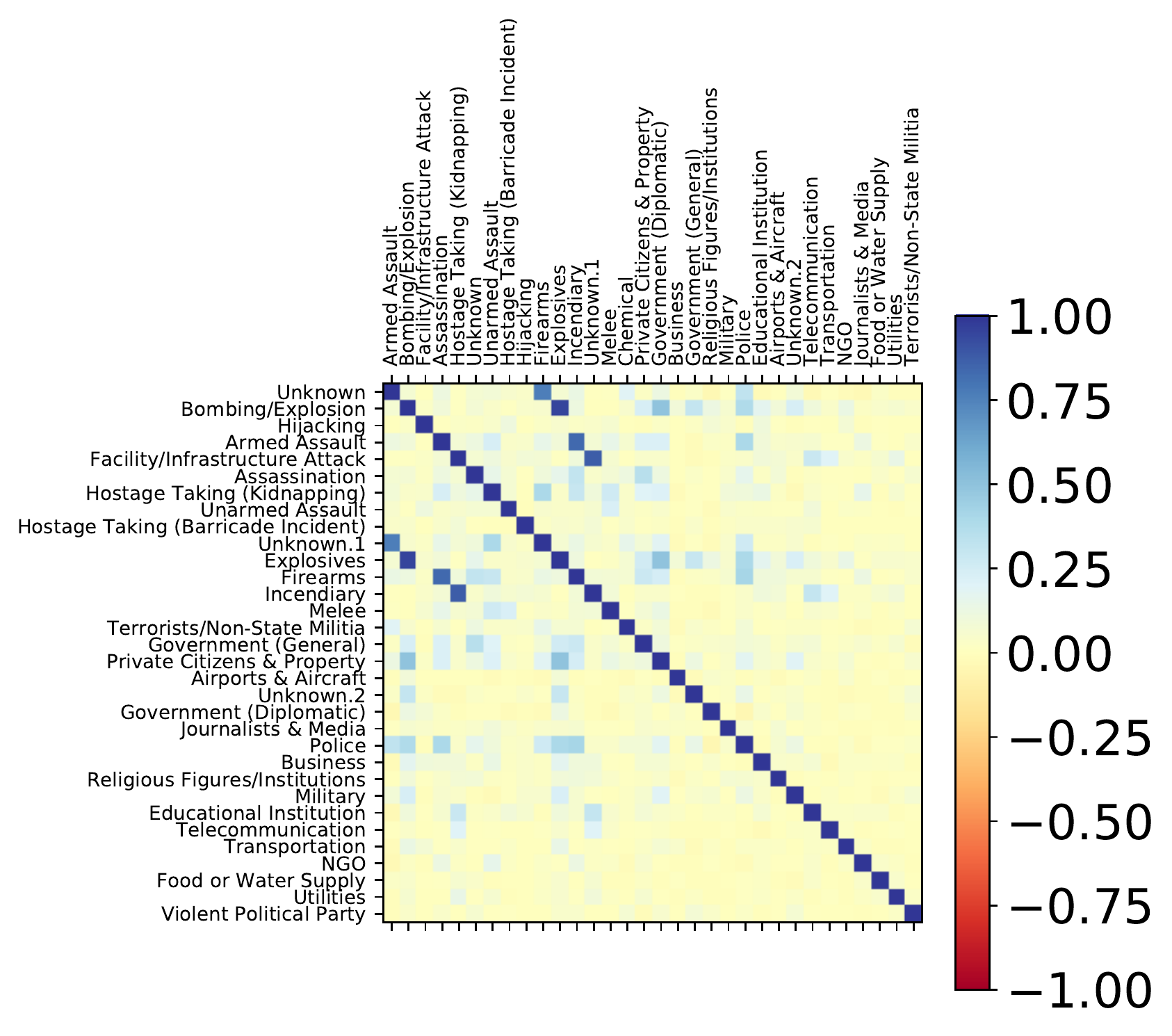} }}%
    \qquad
    \centering
    \subfloat[\centering Iraq]{{\includegraphics[width=8cm]{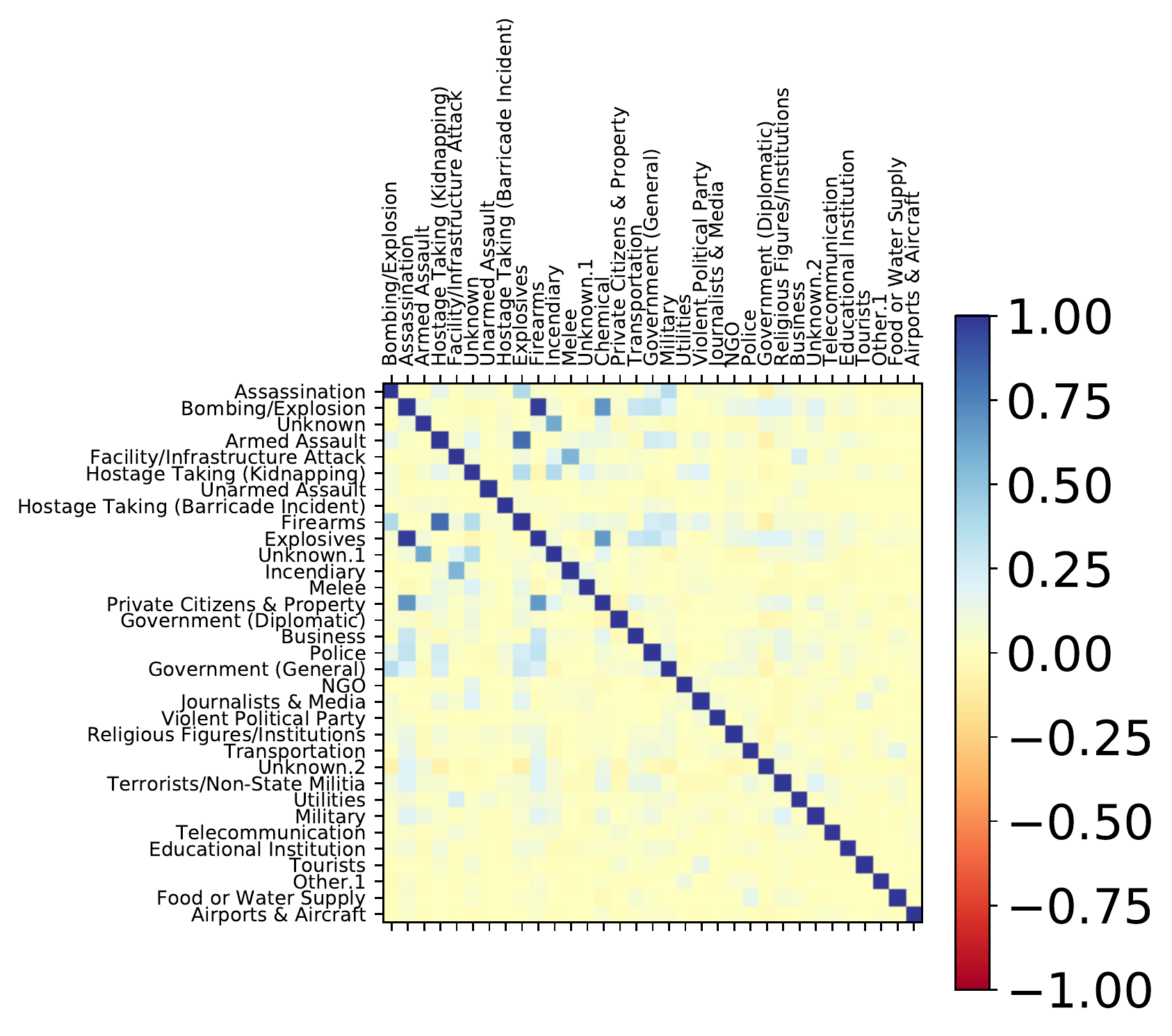} }}%
    \caption{Pearson's correlation of centrality values among all features $F=|X|+|W|+|Y|$ for the entire time-span $U$.}%
    \label{fig:correlation}%
\end{figure}

\begin{figure*}[t]
    \centering
    \subfloat[\centering Afghanistan]{{\includegraphics[scale=.3]{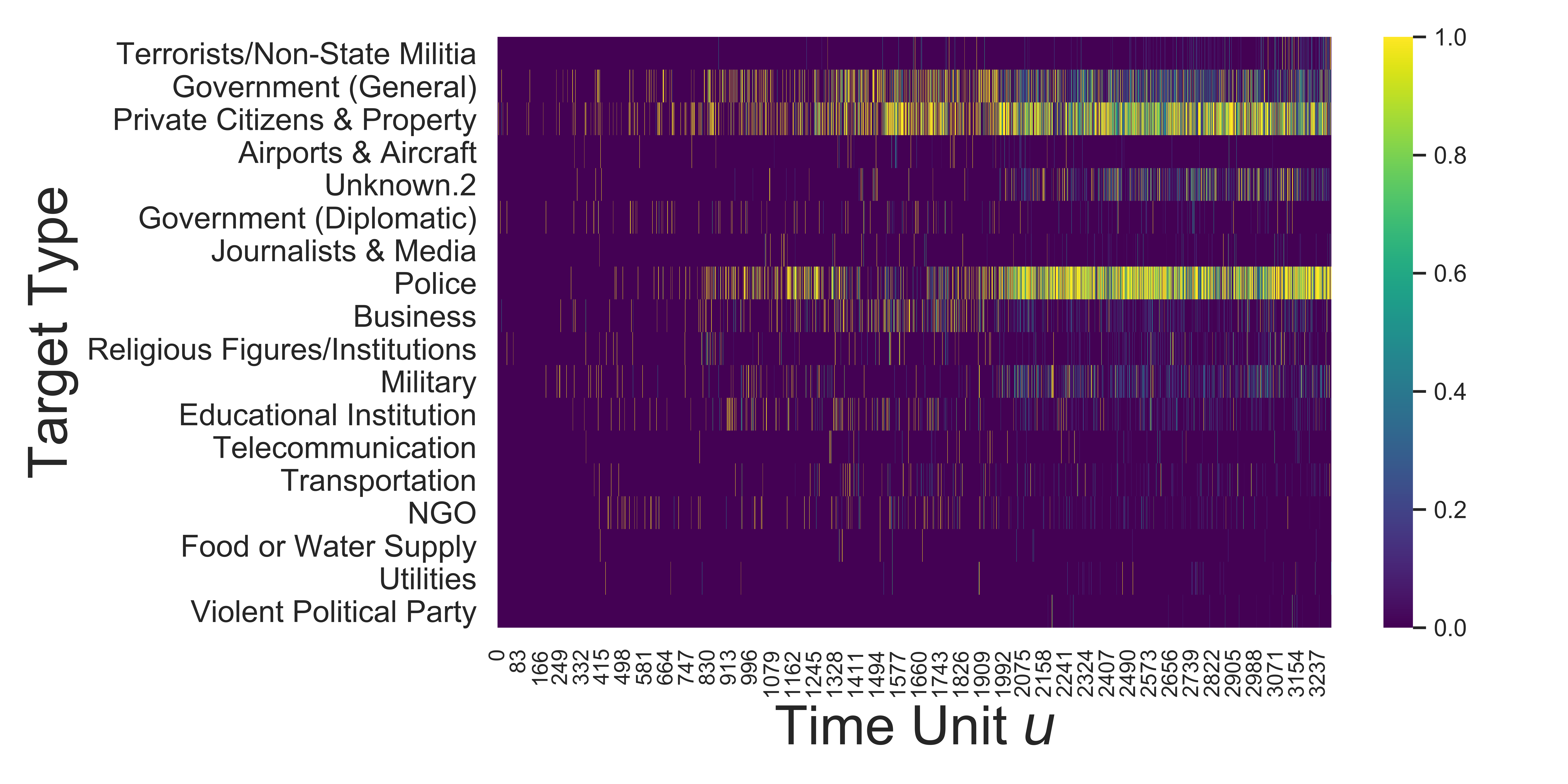} }}%
    \subfloat[\centering Iraq]{{\includegraphics[scale=.3]{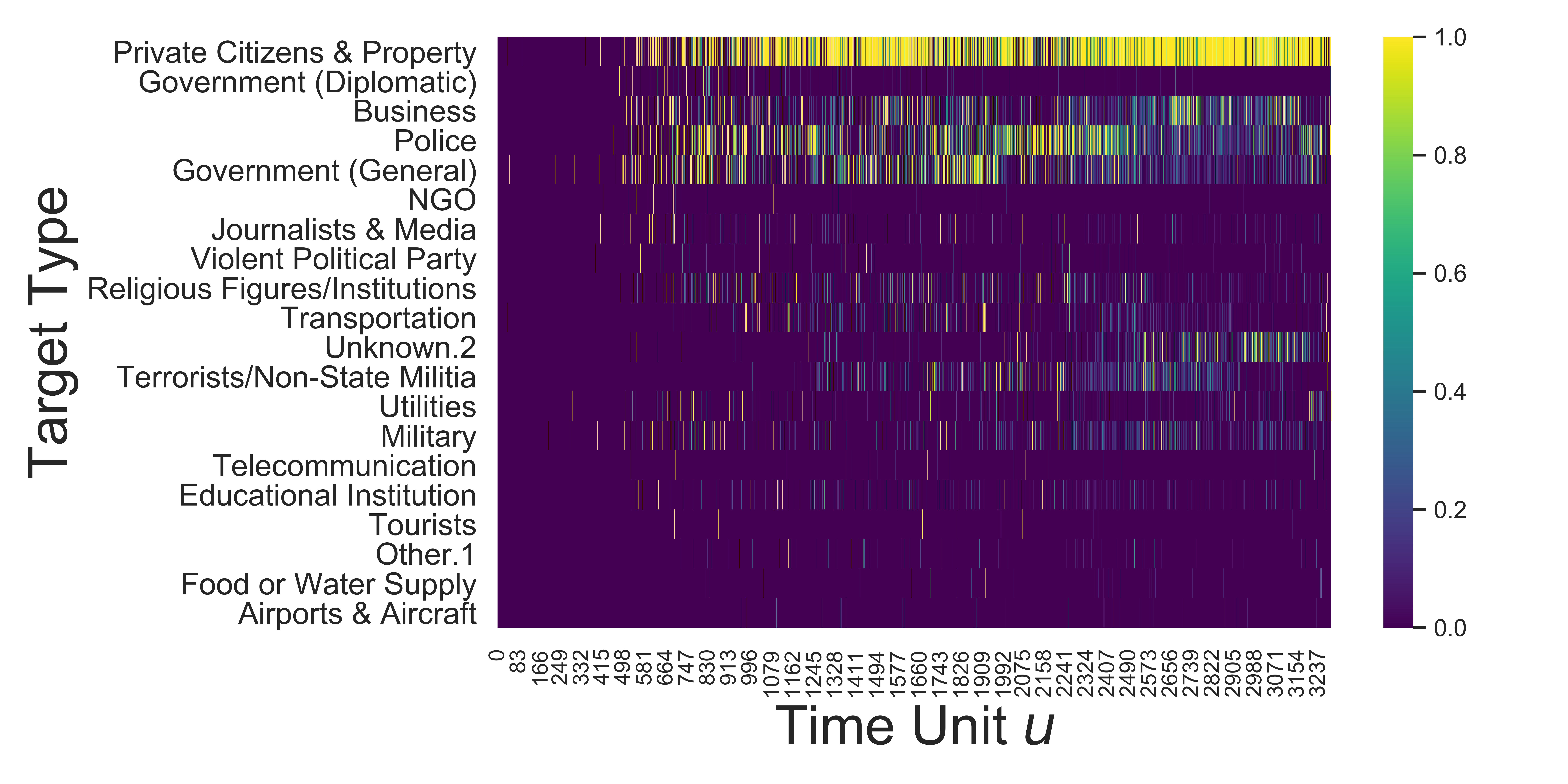} }}%
\caption{Temporal evolution of centrality in the target dimensions  ($\mathcal{D}_{Y}$) for the Afghanistan and Iraq cases.}
    \label{fig:centrality_overtime}
\end{figure*}

Reshaping the data structure from a sequence of graphs to a sequence of continuous values in the range $[0,1]$ simplifies the problem while preserving the relevant relational information emerging from each meta-graph. Figure \ref{fig:correlation} depicts the Pearson's correlation among all features in all dimensions, while the temporal evolution of centrality values for features in the targets dimension $\mathcal{D}_{Y}$ is visualized in Figure \ref{fig:centrality_overtime}.  For both countries, plots describing the characteristics and distributions of tactics, weapons and targets are available in the Supplementary Materials (Figures S1-S4 for Afghanistan, and Figures S5-S8 for Iraq).  

\section{Methods}

\subsection{Algorithmic Setup and Models}
We test the performance of six different types of models on our multivariate time-series forecasting task. Besides the computational contribution of the work in proposing the use of meta-graphs to model terrorist activity over-time, we are also interested in addressing relevant theoretical questions from a terrorism research standpoint. For this reason, the models are trained and fit using different input widths, i.e., a different number of time units. Algorithms performing better with shorter input widths would suggest that terrorist actors rapidly change their strategies and operations, indicating very low memory and thus making it inefficient to rely on large amounts of data. Contrarily, longer input widths would denote that algorithms have to be trained by taking into account relevant portions of recent events' history given a relative degree of stability in attack operations.

In addition, to demonstrate the utility of our meta-graph learning framework, we train the same identical models (both in terms of architectures and input widths) on as many models using a shallow framework that learns target centralities from a feature space engineered simply using aggregate counts of every single weapon and tactics in every time unit. We will then compare the overall results of the two approaches to empirically assess which one better captures the inherent dynamics occurring across terrorist events. 

All the models have been trained using TensorFlow 2.5 \cite{tensorflow2015-whitepaper} and have been initialized with the same random seed. For both datasets, the 70\% of the time units ($u$=2,300; $\sim$ 12.6 years) are used to train the model, the 20\% ($u$= 658, $\sim$ 3.6 years) for validation and the remaining 10\% ($u$=329, $\sim$ 1.8 years) for testing.  Additional modeling details besides the ones presented below are available in the Supplementary Materials in Section S2. 

\paragraph{Baseline.}
The first tested model is a simple baseline scenario in which the forecasted centrality values ${\overline{{\psi}}_{i,\mathrm{Norm}}[u+1]}$ in the next time unit will be equal to the current ground-truth centrality values ${{\psi}_{i,\mathrm{Norm}}[u]}$. This trivial case assumes that terrorist dynamics do not change over time, pointing in the direction of low tendency of terrorist actors to innovate their operational behaviors. 
\paragraph{Feedforward Neural Network.}
The second model tested was a feed-forward dense neural network (FNN) that has two fully connected hidden layers. FNN is the most simple and straightforward class of neural networks and FNNs are not explicitly designed for applications that have time-series or sequence data \cite{Goodfellow-et-al-2016}. In fact, FNNs are not able to capture the temporal interdependencies existing across inputs and thus treat each input sequence as independent of one another. Their inclusion in our experiments is motivated by the interest in investigating whether our multivariate time series actually possess temporal autocorrelations or not through a comparison with other architectures that are designed to handle ordered inputs.
\paragraph{Long Short-Term Memory Network.}
Recurrent Neural Networks (RNN) are one of the algorithmic standards in research problems involving time series data or, more in general, sequences. In the present work, we employ an RNN with Long Short-Term Memory (LSTM) units \cite{HochreiterLongShortTermMemory1997b}. LSTM networks have been proposed to solve the well-known problem of vanishing (or exploding) gradients in traditional RNN. They include memory cells that can maintain information in memory for longer periods, thus allowing the algorithm to more efficiently learn long-term dependencies in the data. We employ LSTM networks with dropout to avoid over-fitting \cite{10.5555/2627435.2670313}, using a two hidden layer structure.
\paragraph{Convolutional Neural Network.}
Convolutional Neural Networks (CNNs) are extensively used in computer vision problems, operating on images and videos for prediction and classification purposes \cite{LeCunDeeplearning2015}. To handle such tasks, CNNs mostly work on 2-dimensional data such as images and videos. However, CNNs can also handle unidimensional signals such as time series \cite{FawazDeeplearningtime2019}. A 1D-CNN incorporates a convolutional hidden layer that applies and slides a filter over the sequence, which can be generally seen as a non-linear transformation of the input. 
\paragraph{Bidirectional LSTM.}
Bidirectional LSTM (Bi-LSTM) are an extension of traditional LSTM models \cite{SchusterBidirectionalrecurrentneural1997}. They train two distinct LSTM layers: the first layer uses the sequence in the traditional forward order, while the other layer is trained on the sequence passed backward. This mechanism allows the network to preserve both information on the past and information on the future, thus fully exploiting the temporal dynamics of the time series under consideration. 
\paragraph{CNN-LSTM.}
To complement the strengths of FNNs, CNNs, and LSTMs, Sainath et al. \cite{SainathConvolutionalLongShortTerm2015} have proposed an architecture combining the three to solve speech recognition tasks. Since then, the approach has demonstrated many promises also in applications involving time-series data. We thus test the performance of a CNN-LSTM inspired by the work of Sainath and colleagues.: the network has a first 1d convolutional layer, followed by a max pooling layer. The information is passed through a dense layer and an LSTM one, respectively. Finally, the network involves an additional dense layer that processes the output.

\subsection*{Performance Evaluation}
To evaluate the performance of the compared algorithms we first use a standard metric, i.e., Mean Squared Error (MSE), computed as:

\begin{equation}
\mathrm{MSE}= \frac{1}{U}\sum_{i=1}^{u}\left ( {{\psi}_{i,\mathrm{Norm}}[u]}-{\hat{{\psi}}_{i,\mathrm{Norm}}[u]} \right )^{2}
\label{eqn:mse}
\end{equation}

It is worth specifying that the evaluation is done on the test data and that in Equations \ref{eqn:mse}-\ref{eqn:swe}, $U$ denotes the length of the test data and $u$ denotes one specific time slice of the test data. We keep this general notation to avoid confusion. We relied on MSE because we aimed at penalizing bigger errors. Furthermore, we then propose two alternative metrics. While centrality values are capturing correlational patterns across different attack dimensions, they are not directly interpretable in real-world terms. For this reason, we shift to a set-perspective and we test the models on their ability to learn the most central targets, i.e. the most popular, frequent, and connected ones, in two different ways.

\paragraph{Element-wise Accuracy}
Element-wise accuracy $\Phi$ (EWA) is the simplest metric between the two. Given the sequence of test data, the sets $S[u]$ and $\hat{S}[u]$ represent the actual set of two most central targets and the predicted set of two most central targets at $u$.  We define the element-wise accuracy $ \phi[u]$  as: 

\begin{equation}
    \phi[u] := \left\{\begin{matrix} 1 & \mathrm{if}\:\: \hat{S}[u] \cap S[u] \neq \varnothing \\ 0 & \mathrm{if}\:\: \hat{S}[u] \cap S[u] = \varnothing \\ \end{matrix}\right.
    \label{six}
\end{equation}
Equation \ref{six} means that if the sets have at least one element in common, then $\phi_{u_{i}}$ is equal to 1, while if the two sets are disjoint the value will be equal to zero. For the entire history of considered time units $U$, then, the overall EW accuracy $\Phi_{U}$ is computed as:

\begin{equation}
    \Phi_{U}= \frac{1}{U}\sum_{i=1}^{U}\phi[u]
\end{equation}
with $\Phi_{U}$ being the ratio between the sum of single unit binary accuracies $\phi[u]$ and the total number of time units $U$.

\paragraph{Set-wise Accuracy}
Set-wise accuracy $\Gamma$ (SWA) is more challenging and further tests the ability of the deep learning models to identify and predict the correct set of targets $S[u]$. The cardinality of $S[u]$ is bounded in the range $0 < \left | S[u] \right |\leq 2$. Thus, for a given time unit $u$, single $\gamma[u]$ is defined as:

\begin{equation}
    \gamma[u]:= \frac{|\hat{S}[u] \cap S[u]|}{|S[u]|}
\end{equation}

In our particular case,  $\gamma[u]$ is equal to 1 if the two sets are perfectly identical (as in any set, it is worth noting, the order does not matter), is 0 when the two sets are disjoint and 0.5 when there is an intersection between $S$ and $\hat{S}$. Finally, the overall metric $\Gamma$ for the sequence $U$ is given by: 

\begin{equation}
   \Gamma_U=\frac{1}{U }\sum_{u=1}^{U}\gamma[u]
   \label{eqn:swe}
\end{equation}
$\Gamma_{U}$ is then simply computed as the average value of all $\gamma[u]$ over the entire sequence of time units $U$. The metric aims at providing more comprehensive information to researchers and intelligence analysts potentially interested in designing logistically wider prevention strategies.

In the analyzed time-series, there is a considerable amount of cases in which time units did not record any attack in both datasets. In Afghanistan, no-attack units account for a 20.54\% of the total time units $u$, while for Iraq this percentage is higher (21.2\%). No-attack units are most prevalent in the first years of the considered time-span, but we needed however to consider this eventuality in the performance evaluation of the proposed models. 

Correctly forecasting no-attack units is crucial. In a counter-terrorism scenario, avoiding forecasting likely targets (and therefore attacks) in time units that do not experience any terrorist action reduces the costs of policy and intelligence strategy deployment. A wrong prediction in a no-attack unit is defined as a case in which: $S[u]= \varnothing \: \: \wedge \: \: \hat{S}[u] \neq \varnothing$, meaning that the set of predicted targets at risk of being hit $\hat{S}[u]$ has at least one target $y$, while the actual set of hit targets $S[u]$ is empty because no attacks were recorded.

Our forecasting framework involves a complex forecasting task. It is complex for two main interconnected reasons. First, we train our models to predict a relatively large set of time series characterized by sparsity and non-stationarity. Second, the number of time units is limited. Deep learning for time series prediction or classification has been deployed in many domains, including high-frequency finance applications \citep{Baodeeplearningframework2017, SezerFinancialtimeseries2020} or weather and climate prediction \citep{SHIConvolutionalLSTMNetwork2015, QingHourlydayaheadsolar2018, AlemanyPredictingHurricaneTrajectories2019, WangOMuLeTOnlineMultiLead2020}. However, unlike these forecasting settings, open-access data on terrorist events cannot be measured using micro-scales such as minutes, seconds or hours, because data are collected at the daily level, hence leading to a significantly limited number of two-day based time units $u$. 

Therefore, the models have to quickly learn complex temporal and operational inter-dependencies without relying on massive amounts of data. Given these premises, it is highly unlikely that the forecasts can exactly predict centrality values $\psi_{i_{\mathrm{Norm}}}[u]$ that are equal to 0 for each one of the features in the $X$, $W$ and $Y$ dimensions. Forecasts will contain noise, leading to very small centrality estimates for a number of features. When those very small $\psi_{i_{\mathrm{Norm}}}[u]$ are produced as estimates, the element-wise and set-wise accuracy metrics will produce erroneous results.
To accommodate the necessity to correctly forecast no-attack days, we have set defined a rule that overcome the likely presence of noise-driven values. Given $\ddot{u}$, that is a time unit $u$ with zero attacks, we have thus set a threshold $\xi=0.1$, such that: 


\begin{equation}
\phi [\ddot{u}], \gamma[\ddot{u}] := \left\{\begin{matrix} 1 & \hat{\psi}_{i,\mathrm{Norm}}<\xi , \forall i \in I \\ 0 & \mathrm{otherwise} \end{matrix}\right.
\label{noattackeq}
\end{equation}

Equation \ref{noattackeq} means that the element wise accuracy ${\phi}[\ddot{u}]$ and the set-wise accuracy $\gamma [\ddot{u}]$ of the time unit $\ddot{u}$ are both equal to 1 if all the predicted centrality values $\hat{\psi}_{i,{\mathrm{Norm}}}[\ddot{u}]$ are below the threshold $\xi=0.1$. 

\begin{figure}[t]
    \centering
    \subfloat[\centering Afghanistan]{{\includegraphics[scale=.4]{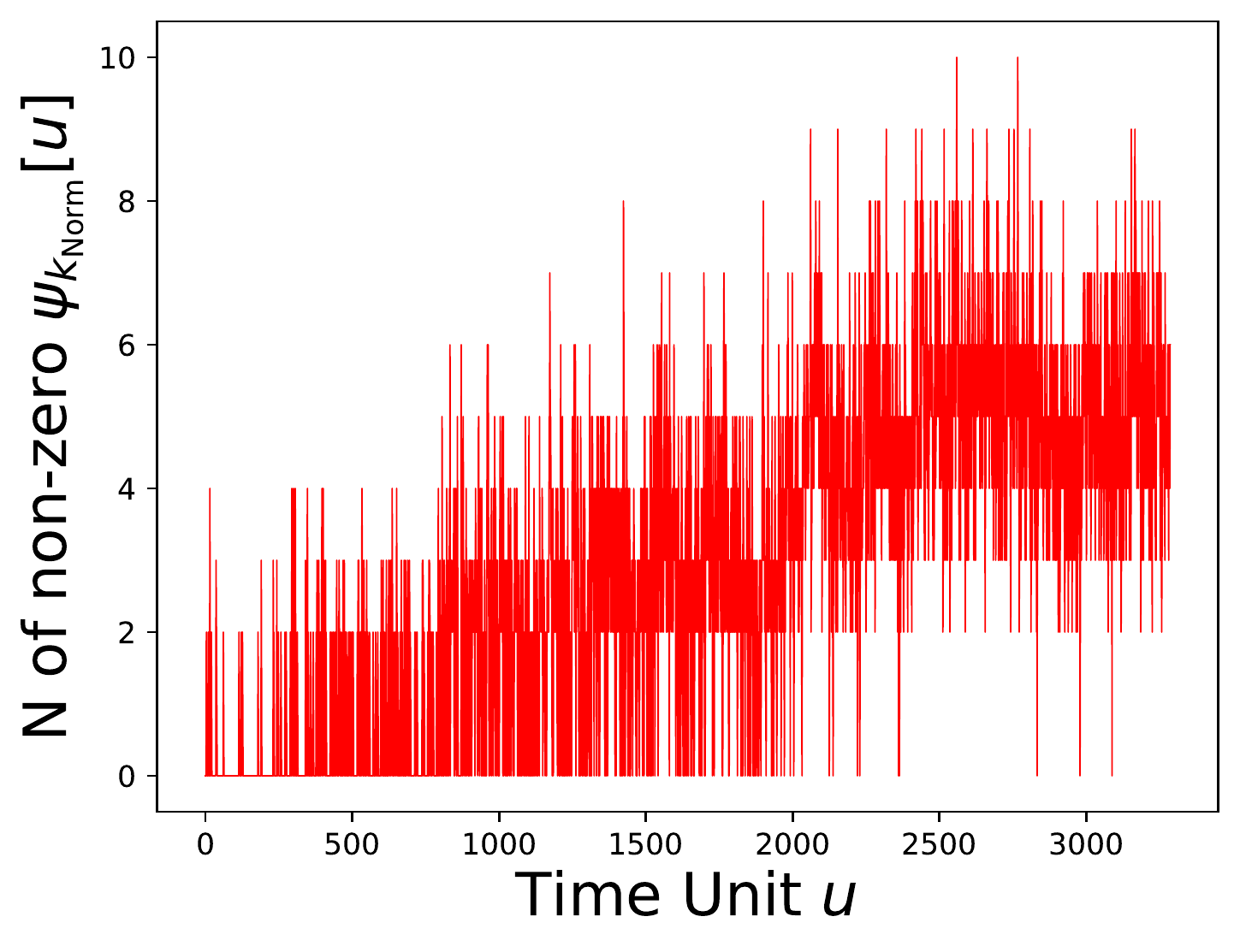} }}%
    \qquad
    \subfloat[\centering Iraq]{{\includegraphics[scale=.4]{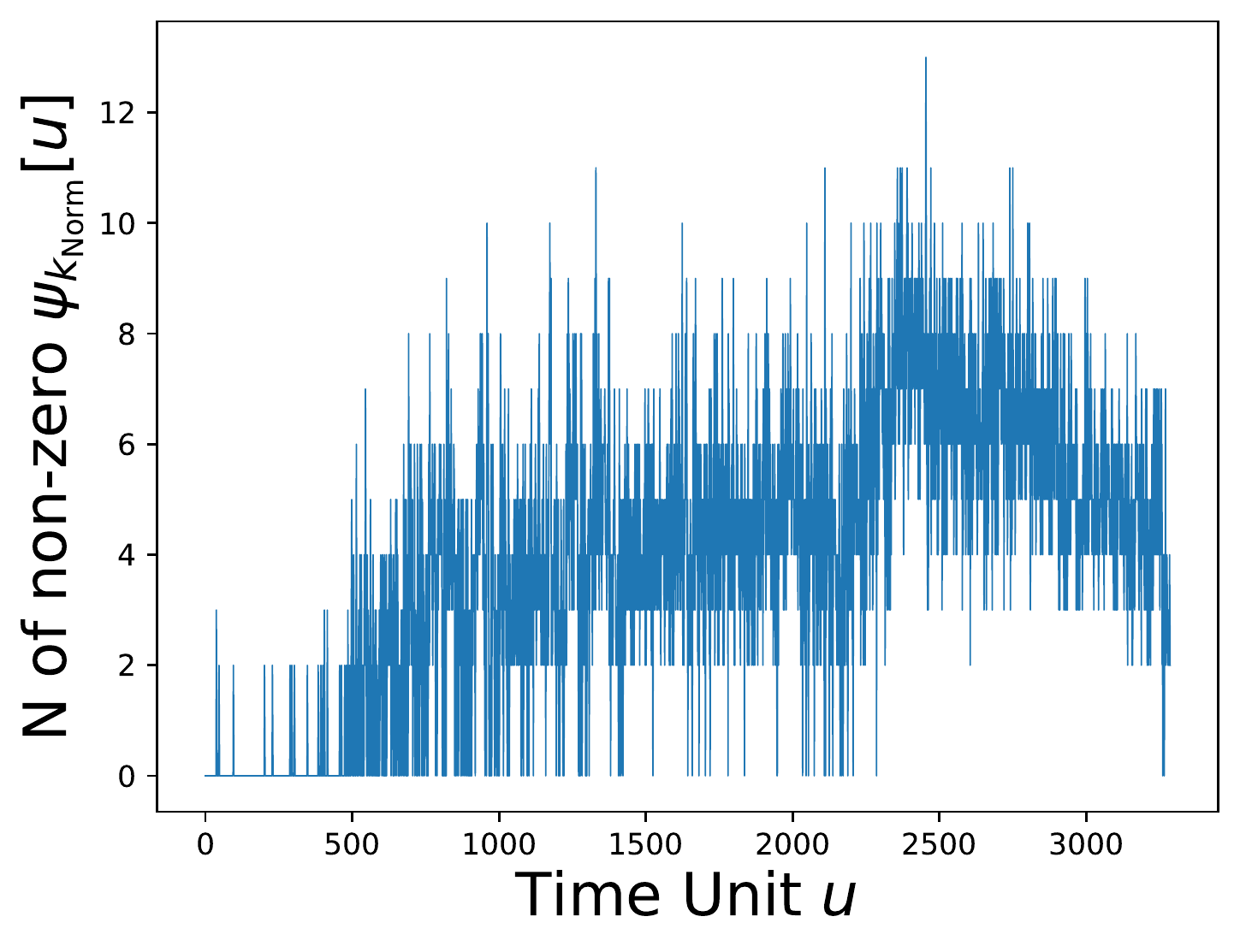} }}%
\caption{Temporal Trend of the count of non-zero centrality values $ \psi_{k_{\mathrm{Norm}}}[u]$ in the target dimension for Afghanistan and Iraq.}
    \label{fig:centrality_nonzero}
\end{figure}

Conversely, if  $\exists i$ s.t. $\hat{\psi}_{i,{\mathrm{Norm}}}[\ddot{u}]>\xi$, then the evaluation of both the metrics is equal to 0.  Given the distributional ranges of all the values (see Supplementary Materials), the $\xi$ threshold set at 0.1 resulted in the right compromise between an excessive penalization of the algorithm performance and a shallow assessment of the actual forecasting capabilities of the proposed framework. Figure \ref{fig:centrality_nonzero} shows the temporal trend of counts of non-zero centrality values in the target dimension for Afghanistan and Iraq.

\section{Results}

\begin{table*}[!hbt]
\setlength{\tabcolsep}{3.5pt}
\footnotesize
\centering
\begin{tabular}{l|c||ccc|ccc||ccc|ccc}
\hline
\multirow{3}{*}{\textbf{Model}} & \multirow{3}{*}{\textbf{\begin{tabular}[c]{@{}c@{}}Input\\ Width\end{tabular}}} & \multicolumn{6}{c||}{\textbf{META-GRAPH LEARNING}} & \multicolumn{6}{c}{\textbf{SHALLOW LEARNING}} \\ \cline{3-14} 
 &  & \multicolumn{3}{c|}{\textbf{Afghanistan}} & \multicolumn{3}{c||}{\textbf{Iraq}} & \multicolumn{3}{c|}{\textbf{Afghanistan}} & \multicolumn{3}{c}{\textbf{Iraq}} \\ \cline{3-14} 
 &  & \textbf{$\Gamma$} & \textbf{$\Phi$} & \textbf{MSE} & \textbf{$\Gamma$} & \textbf{$\Phi$} & \textbf{MSE} & \textbf{$\Gamma$} & \textbf{$\Phi$} & \textbf{MSE} & \textbf{$\Gamma$} & \textbf{$\Phi$} & \textbf{MSE} \\ \hline
Baseline & 1 & 0.1570 & 0.3140 & 0.7733 & 0.0701 & 0.1371 & 0.2371 & 0.1570 & 0.3140 & 0.7733 & 0.0701 & 0.1371 & 0.2371 \\
FNN & 1 & 0.6478 & 0.9146 & 0.0399 & 0.5594 & 0.9176 & 0.0275 & 0.6036 & 0.8658 & 0.0448 & 0.5183 & 0.8841 & 0.0527 \\
LSTM & 1 & 0.6753 & 0.9237 & 0.0395 & 0.5716 & 0.9237 & 0.0272 & 0.6265 & 0.8963 & 0.0410 & 0.5320 & 0.9024 & 0.0281 \\
CNN & 1 & 0.6204 & 0.8780 & 0.0414 & 0.5564 & 0.9176 & 0.0266 & 0.5807 & 0.8658 & 0.0485 & 0.5182 & 0.8963 & 0.0787 \\
Bi-LSTM & 1 & 0.6692 & 0.9238 & 0.0372 & 0.5442 & 0.9112 & 0.0261 & 0.6265 & 0.8871 & 0.0406 & 0.4969 & 0.8963 & 0.0268 \\
CNN-LSTM & 1 & 0.6494 & 0.9116 & 0.0393 & 0.5457 & 0.9146 & 0.0269 & 0.6189 & 0.8871 & 0.0382 & 0.5259 & 0.8993 & 0.0269 \\ \hline
FNN & 5 & 0.6204 & 0.8920 & 0.0415 & 0.5663 & 0.9228 & 0.0261 & 0.5432 & 0.8333 & 0.0569 & 0.5216 & 0.9012 & 0.0486 \\
LSTM & 5 & 0.6651 & 0.9136 & 0.0392 & 0.5771 & 0.9290 & 0.0254 & 0.5633 & 0.8395 & 0.0407 & 0.5370 & 0.9135 & 0.0248 \\
CNN & 5 & 0.6404 & 0.8889 & 0.0395 & 0.5524 & 0.9105 & 0.0265 & 0.5462 & 0.8364 & 0.0591 & 0.5000 & 0.8858 & 0.0541 \\
Bi-LSTM & 5 & 0.6836 & \textbf{0.9352} & \textbf{0.0366} & \textbf{0.5787} & \textbf{0.9290} & 0.0246 & 0.6404 & 0.9104 & 0.0370 & 0.5401 & 0.9104 & \textbf{0.0238} \\
CNN-LSTM & 5 & 0.6327 & 0.8920 & 0.0403 & 0.5370 & 0.8920 & 0.8920 & 0.5802 & 0.8488 & 0.0389 & 0.5201 & 0.8981 & 0.0249 \\ \hline
FNN & 15 & 0.6210 & 0.8885 & 0.0442 & 0.5478 & 0.8949 & 0.0286 & 0.6369 & 0.9044 & 0.0450 & 0.5207 & 0.8949 & 0.0567 \\
LSTM & 15 & 0.6768 & 0.9236 & 0.0376 & 0.5732 & 0.9267 & 0.0263 & 0.6257 & 0.8949 & 0.0370 & 0.5286 & 0.9012 & 0.0255 \\
CNN & 15 & 0.6322 & 0.8949 & 0.0441 & 0.5478 & 0.8980 & 0.0317 & 0.5859 & 0.8630 & 0.0503 & 0.5254 & 0.8949 & 0.0531 \\
Bi-LSTM & 15 & 0.6863 & 0.9331 & 0.0370 & 0.5573 & 0.9172 & 0.0240 & 0.6496 & 0.9171 & \textbf{0.0366} & 0.5000 & 0.8949 & 0.0248 \\
CNN-LSTM & 15 & 0.6433 & 0.8949 & 0.0440 & 0.5398 & 0.8917 & 0.0246 & 0.6114 & 0.8757 & 0.0384 & 0.4952 & 0.8789 & 0.0265 \\ \hline
FNN & 30 & 0.5335 & 0.8361 & 0.0520 & 0.5669 & 0.9163 & 0.0311 & 0.5785 & 0.8127 & 0.0485 & 0.5301 & 0.8996 & 0.0429 \\
LSTM & 30 & 0.6756 & 0.9298 & 0.0377 & 0.5735 & 0.9231 & 0.0263 & 0.6054 & 0.8795 & 0.0431 & 0.5301 & 0.9063 & 0.0250 \\
CNN & 30 & 0.6187 & 0.8829 & 0.0478 & 0.5619 & 0.9130 & 0.0331 & 0.5619 & 0.8361 & 0.0592 & 0.5317 & 0.8963 & 0.0451 \\
Bi-LSTM & 30 & \textbf{0.6890} & 0.9333 & 0.0369 & 0.5769 & 0.9230 & 0.0268 & 0.6572 & 0.9130 & 0.0376 & 0.5017 & 0.8862 & 0.0252 \\
CNN-LSTM & 30 & 0.5852 & 0.8729 & 0.0482 & 0.5585 & 0.9164 & 0.0250 & 0.5953 & 0.8662 & 0.0452 & 0.5183 & 0.8996 & 0.0301 \\ \hline
\end{tabular}
\caption{Performance in terms of Set-wise Accuracy ($\Gamma$), Event-wise Accuracy ($\Phi$) and Mean Squared Error (MSE) across chosen algorithms with varying input widths. For $\Gamma$ and $\Phi$, higher values mean better performance. The contrary holds for MSE. Values in bold indicate the best performance obtained in each dataset overall (i.e., taking into consideration also Shallow datasets).}
\label{results2}
\end{table*}

\begin{figure}[hbt!]%
    \centering
    \subfloat[\centering Afghanistan]{{\includegraphics[width=8cm]{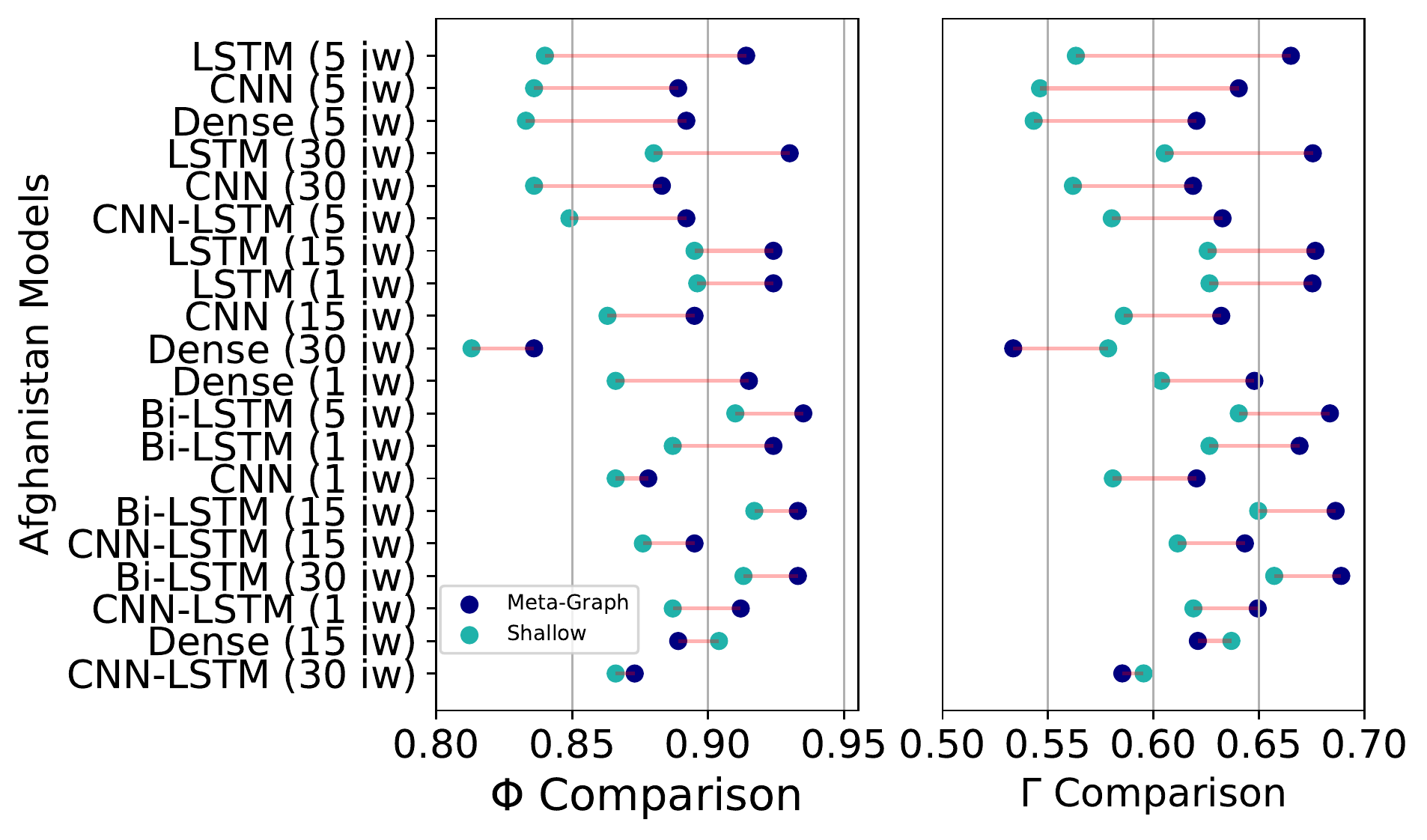} }}%
    \qquad
    \subfloat[\centering Iraq]{{\includegraphics[width=7.8cm]{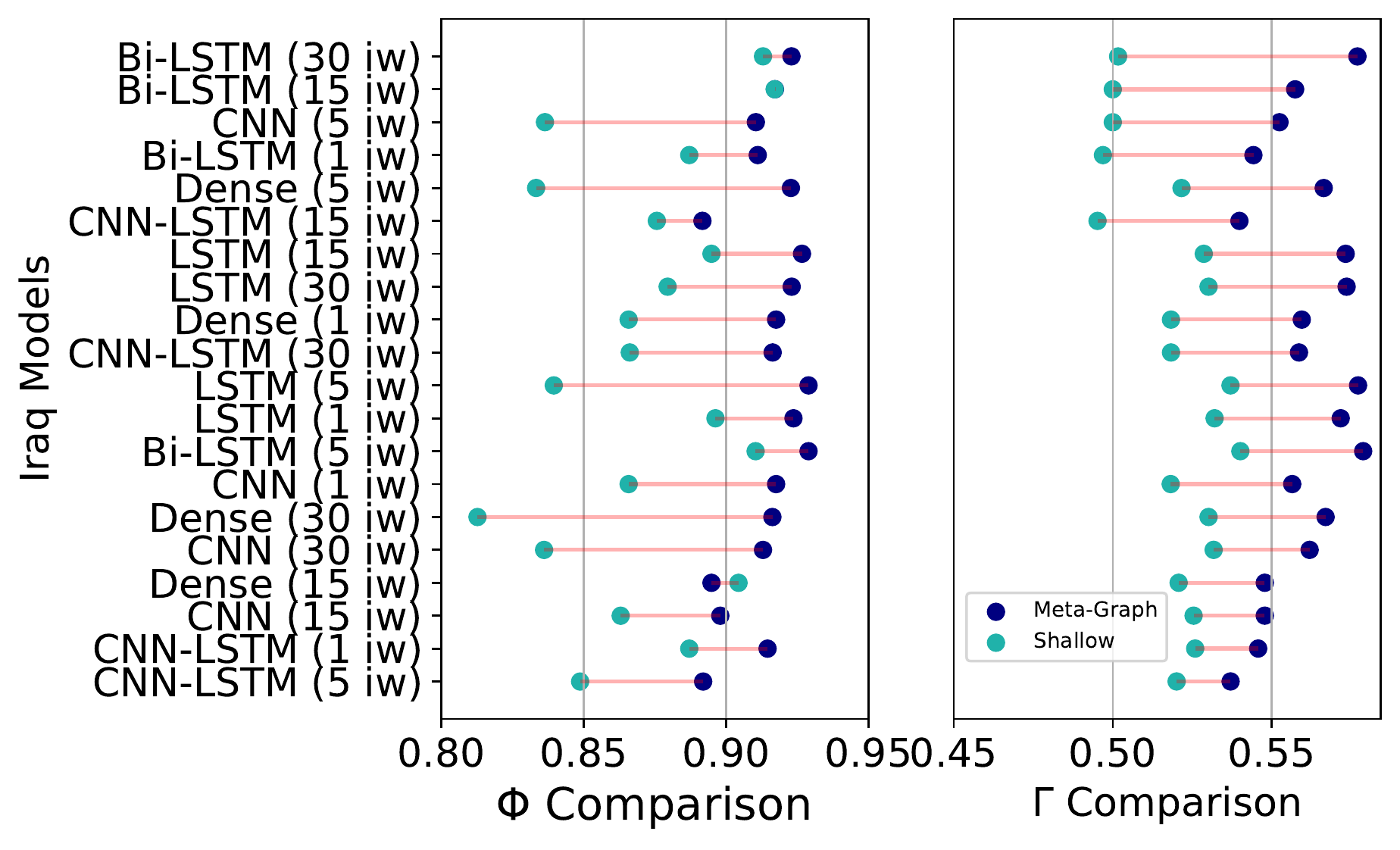} }}%
    \caption{Comparison between Meta-Graph and Shallow Learning for Afghanistan and Iraq. Models are ordered based on descending difference in $\Gamma$ (Set-wise Accuracy) performance.  ``IW'' indicates  ``input width''.}%
    \label{fig:compa}%
\end{figure}

The results reported in Table \ref{results2} overall showcase two relevant sets of results. First, the models clearly illustrate that structuring the problem of forecasting the next most central terrorist targets through a meta-graph learning framework outperforms an alternative scenario in which targets are forecasted simply using the shallow count of feature occurrences. This suggests that engineering the feature space by exploiting inter-dependencies among events in a network fashion allows capturing more information regarding the patterned nature of terrorist activity. In both the Afghanistan and Iraq cases, $\Gamma$ and $\Phi$ are always higher when using graph-derived time-series. One exception is given by the MSE, which appeared to be lowest in the shallow-learning case for the Iraq dataset and \textit{ex-aequo} in the Afghanistan one. While a shallow learning approach works reasonably well in terms of regression, the meta-graph scenario works consistently better when forecasted targets need to be correctly ranked in terms of their ground-truth centrality. The comparison between meta-graph learning and shallow learning performances is provided in Figure \ref{fig:compa}. 

Second, the outcomes indicate that Bi-LSTM models always reach higher results in terms of $\Phi$ and $\Gamma$, regardless of the considered dataset and feature engineering approach. This convergence demonstrates that fitting a model that can rely on both backward and forward inputs enables the neural network to gather much richer contextual information, leading to superior forecasting performance. 

Third, the outcomes also reveal that a baseline model with no learning component assuming no changes in terrorist strategies is highly inaccurate and, furthermore, the performance of Feedforward Dense Networks underscores that treating centrality sequences without taking into consideration the underlying temporal connections across different time units leads to sub-optimal predictions

We were also interested in assessing the model results not only across algorithmic architectures but also across different input widths, contributing both to the computational and the theoretical study of terrorism. Highly similar results emerged, with a notable exception. A Bi-LSTM model trained using 5 time units $u$ as input widths have reached the highest performance in both datasets for what concerns $\Phi$ and in the Iraq case also in terms of $\Gamma$. However, in the Afghanistan dataset, the same model architecture using 30 time-units $u$ as input width obtained the highest $\Gamma$. The need for six times more data points for obtaining optimal forecasts denotes the higher stability of terrorism in Afghanistan, and a more microcycle-like structure of events in Iraq \citep{BehlendorfMicrocyclesViolenceEvidence2012}. This difference (although minimal) between the Afghanistan and Iraq cases might be explained by a higher heterogeneity of strategies and, ultimately, actors involved and opens new lines of inquiry to understand what factors are driving these dynamics (e.g., higher creativity, more resources, internal fighting).

\begin{figure}[!h]%
    \centering
    \subfloat[\centering Afghanistan]{{\includegraphics[scale=0.3]{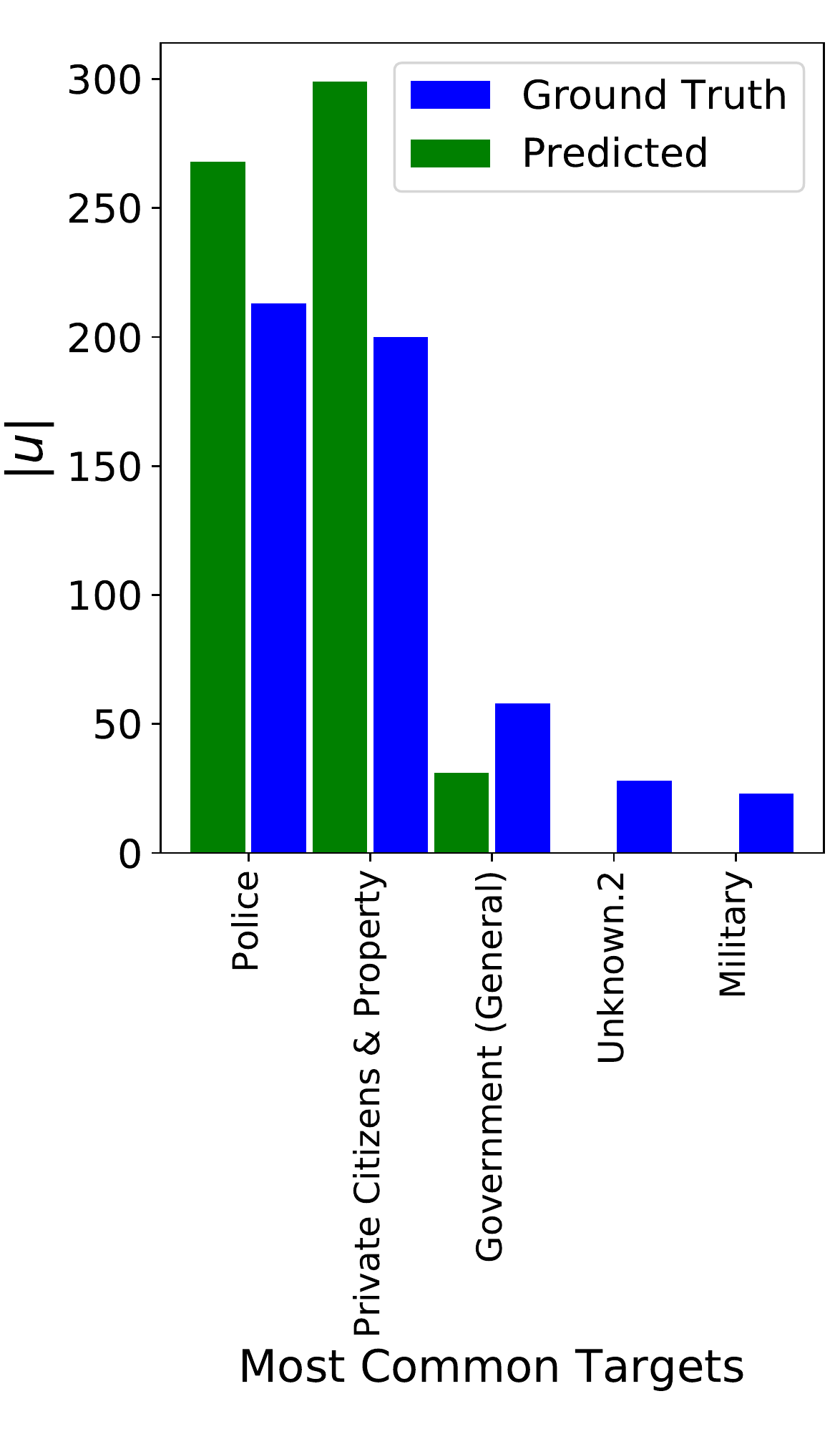} }}%
    \qquad
    \subfloat[\centering Iraq]{{\includegraphics[scale=0.3]{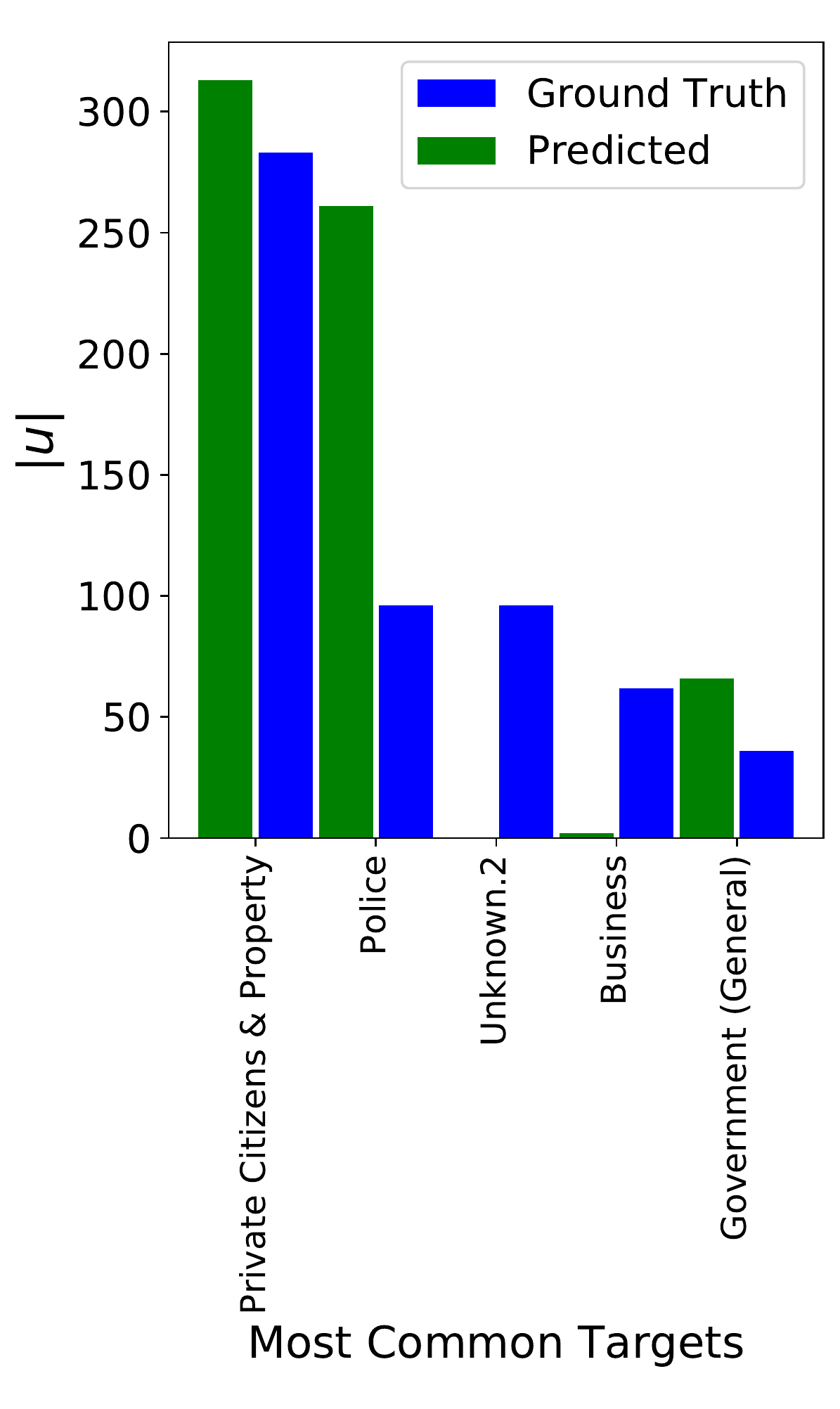} }}%
    \caption{Bar chart comparing the number of time units $|u|$ a feature was empirically ranked among the two highest centrality values in the test set and number of times the same feature was forecasted among the two highest centrality values. The graph reports the result for the model with the highest $\Gamma$ for Afghanistan and Iraq, respectively (Afghanistan: Bi-LSTM with 30 as input width; Iraq: Bi-LSTM with 5 as input width) and shows the five most common targets in each dataset.}%
    \label{fig:toptop}%
\end{figure}
To further inspect the quality and characteristics of the forecasting models, Figure \ref{fig:toptop} conveys target-level information for both datasets. Specifically, the bar charts report a comparison between the empirical number of time units $|u|$ in which a certain target was among the highest two centrality values and the corresponding predicted number of time units in which the same target was forecasted to be in the top-two set, limiting the comparison to the outcomes of the model achieving the best $\Gamma$ for each country. For Afghanistan and Iraq, Figures S9 and S10 in the Supplementary Materials also show the correlation between each vector $\Psi_{norm}[u]$ and $\hat{\Psi}_{norm}[u]$, representing respectively the empirical centrality values of each target feature at each time stamp $u$ and the corresponding predicted centrality values.

One common feature appears for both datasets: the Bi-LSTM models fail when dealing with time unit vectors having Unknown.2 as highly relevant target type (``Unknown.2'' is used to distinguish unknown targets from unknown tactics, ``Unknown'', and unknown weapons, ``Unknown.1'', in the dataset). This is probably due to the peculiar mix of patterns associated with this specific target feature. Additionally, in Afghanistan the model poorly performs in predicting a central role for Military targets and in Iraq the Bi-LSTM model reaches unsatisfactory results when Business appears to be empirically relevant. 

It is worth noting how the Afghanistan distribution of the top five empirical targets appears slightly more unbalanced compared to the Iraq one, possibly influencing the overall model results commented in Table \ref{results2}, which are generally higher in Afghanistan. In both case studies, the forecasting models overestimate the prevalence of the most important empirical targets, calling for future efforts to engineer models that are more capable of capturing the nuances hidden under the different inter-connected dimensions of terrorist actions. Improving the model ability to handle these nuances will lead to a higher propensity of disentangling anomalous combinations of tactics, weapons and targets that may be hard to learn for the algorithms in their present form (possibly due to the low amount of data used compared to standard deep learning applications and the underlying non-stationarity of certain signals used in the datasets). This, in turn, would lead to forecasting distributions increasingly resembling empirical ones, which is the ultimate aim as results could be meaningfully used and deployed also in reference to rare events, intended as the outcomes of rare operational combinations.

In spite of these limits, the satisfactory performance recorded for the two best models (Afghanistan: $\Gamma=0.6890$; Iraq: $\Gamma=0.5787$) along with the visualized distribution of the most central targets corroborates the position of Clarke and Newman \cite{ClarkeOutsmartingTerrorists2006} who repeatedly claimed the importance of protecting a restricted number of possible terrorist targets as a meaningful way to counter and prevent terrorist violence. While, in principle, terrorists have an almost infinite number of possible targets to choose from, their choice is limited by a number of constraints and motivations (both material and immaterial): this translates into the shrinking of the options' spectrum, and in the recurring consistency of a very limited number of target types. In line with Clarke and Newman's argument, the presented models underscore how suboptimal models can still provide effective intelligence knowledge given the patterned nature of terrorist actors, a mixed effect of the strategic character of their actions and the bounded portfolio of resources and options at their disposal.

\section{Discussion and Conclusions}
Artificial Intelligence and computational approaches are increasingly gaining momentum in the study of societal problems, including crime and terrorism. To contribute to this developing area of research, this paper proposed a novel computational framework designed to investigate terrorism dynamics and forecast future terrorist targets. Relying on real-world data gathered from the GTD, we presented a method for representing the complexity and interdependence of terrorist attacks that  relied on the extraction of temporal meta-graphs from thousands of events occurred from 2001 to 2018.  We further tested the ability of six different modeling architectures to correctly predict the targets to be at the most risk of being chosen in the next two days. The work presented the outcomes of multiple experiments performed focusing on terrorist activity in Afghanistan and Iraq. We first compared our approach using temporal meta-graphs against a shallow learning scenario in which forecasts are obtained using a feature space that only considers the count of occurrences across operational features of terrorist events. The comparison demonstrated that using temporal meta-graphs to construct time-series leads to superior forecasting performance, suggesting that embedding event inter-dependencies into temporal sequences offers richer context and information to capture terrorist complexity.

Additionally, results hint that Bidirectional LSTM (Bi-LSTM) models outperform the other architectures in both datasets, although differences arise across the two in terms of forecasting performance and optimal input width. 

This work addressed an unexplored research problem in the growing literature that applies artificial intelligence for social impact, highlighting the potential of machine and deep learning solutions in forecasting terrorist strategies. These promising results call for future research endeavors that should address some of the limitations of the current work, including the generalizability of the results in other geographical contexts, the discrimination between actors operating in the same country, and the use of alternative contextual information such as data on military campaigns or civil conflicts.  In terms of limitations, we specifically highlight that our approach does not take into account geo-spatial information. This decision is justified by the assumption that events occurring in the same country are part of a high-level strategic decision-making process, and that this high-level process is decoded through unified event dynamics. While this assumption is more easily justifiable in countries where one or very few terrorist actors are active or where, in spite of large pools of active actors most events are associated with one or few of them (e.g., Afghanistan), we recognize that it becomes more problematic for countries experiencing activity from a higher number of groups or organizations, as it is the case for Iraq. Patterns that exist at a national level may hide existing dynamics at the regional or provincial level. Future work should thus consider a localized geographical dimension to provide forecasts that have a deeper practical value, as they would produce heterogeneous probabilistic risk scales for different territories, helping intelligence agencies in setting up more tailored counter-terrorism strategies.

\section*{Availability of materials and data}
The data and code used to conduct the analyses here presented are made available at the following GitHub repository: \href{https://github.com/gcampede/terrorism-metagraphs}{https://github.com/gcampede/terrorism-metagraphs}

\section*{Acknowledgments}
This work was supported by the University of Trento - Department of Excellence initiative funded by the Italian Ministry of University and Research and in part by the Knight Foundation and the Office of Naval Research grant N000141712675.
Additional support was provided by the Center for Computational Analysis of Social and Organizational Systems
(CASOS) and the Center for Informed Democracy and Social Cybersecurity (IDeaS). The views and conclusions contained in this document are those of the authors and should not be interpreted as representing the
official policies, either expressed or implied, of the Italian Ministry of University and Research, the Knight Foundation, Office of Naval Research or the U.S. government.

\newpage

\appendix

\setcounter{equation}{0}
\setcounter{section}{0}
\setcounter{figure}{0}
\setcounter{table}{0}
\makeatletter
\renewcommand{\theequation}{S\arabic{equation}}
\renewcommand{\thefigure}{S\arabic{figure}}


\section{Datasets: Descriptive Statistics}
\subsection{Preliminary Filtering}
According to the Global Terrorism Database (GTD) \citep{LaFreeIntroducingGlobalTerrorism2007}, between 2000 and 2018, Afghanistan and Iraq recorded a total of 14,385 and 25,896 terrorist attacks, respectively. As described in the main manuscript, two levels of criteria are imposed for the inclusion of an event in the GTD. Besides these two levels, an additional variable is added to the dataset mapping those events for which doubt exist regarding their terrorist nature. 

In order to avoid biases and noise in our signals, during the generation of the proposed meta-graphs and the time series, we proceeded to exclude all those events that were doubtful in terrorist nature, according to the \textit{doubtterr} variable. This led to a slight reduction in the total number of attacks: 12,120 attacks for Afghanistan and 22,773 for Iraq.

The next two subsections will provide an overview of the descriptive statistics of the temporal meta-graph derived multivariate time series for both Afghanistan and Iraq.

\subsection{Afghanistan}
\begin{figure}[!h]
    \centering
    \includegraphics[scale=0.35]{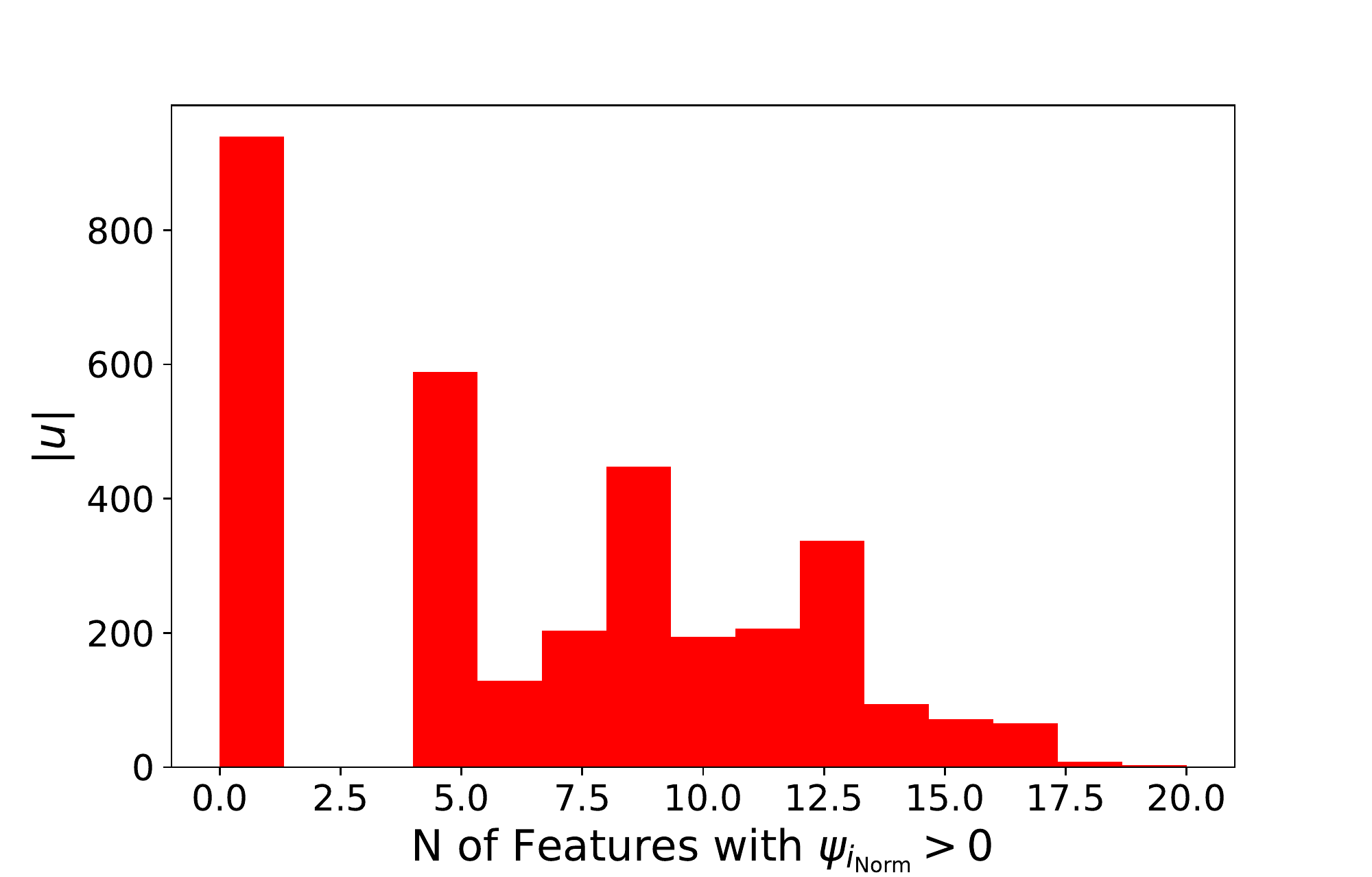}
    \caption{Count of non-zero $\psi_{i_{\mathrm{Norm}}}[u]$ in each $u$ - Afghanistan.}
    \label{fig:nonzero_afg}
\end{figure}

The Afghanistan multivariate time-series, derived from our framework, focus on two-day time units, for a total of 3,289 data points (the same for Iraq). In the period under consideration,  939 units did not record attacks, a 28.54\% of the total. The histogram below (Figure \ref{fig:nonzero_afg}) reports the count of non-zero features in the considered time series, for each time unit (i,e., how many weapons, tactics and targets have a centrality value higher than 0 at time unit $u$?).

\subsubsection{Tactics}
Terrorist actors that were active in Afghanistan between 2000-2018 have exploited 9 different tactic types: \textit{Unknown, Bombing/Explosion,	Hijacking,	Armed Assault,	Facility/Infrastructure Attack,	Assassination,	Hostage Taking (Kidnapping), Unarmed Assault, Hostage Taking (Barricade Incident)}. Below are reported the number of occurrences of each tactic type (Table \ref{afg_tac_tab}). 

\begin{table}[!hbt]
\centering
\footnotesize
\begin{tabular}{l c c}
\hline
\textbf{Tactic $x_i$} & \textbf{N of $\psi_{i_{\mathrm{Norm}}}[u]>0$} & \textbf{\%} \\ \hline
Bombing/Explosion & 1961 & 59.62\% \\ \hline
Armed Assault & 1442 & 43.84\% \\ \hline
Hostage Taking (Kidnapping) & 808 & 24.57\% \\ \hline
Assassination & 781 & 23.75\% \\ \hline
Unknown & 606 & 18.43\% \\ \hline
Facility/Infrastructure Attack & 335 & 10.19\% \\ \hline
Unarmed Assault & 59 & 1.79\% \\ \hline
Hostage Taking (Barricade Incident) & 36 & 1.09\% \\ \hline
Hijacking & 18 & 0.55\% \\ \hline
\end{tabular}
\caption{Afghanistan Tactics - Number of non-zero $\psi_{i_{\mathrm{Norm}}}[u]$  occurrences and percentage over $U$}
\label{afg_tac_tab}
\end{table}

\begin{figure}[!hbt]
    \centering
    \includegraphics[scale=0.10]{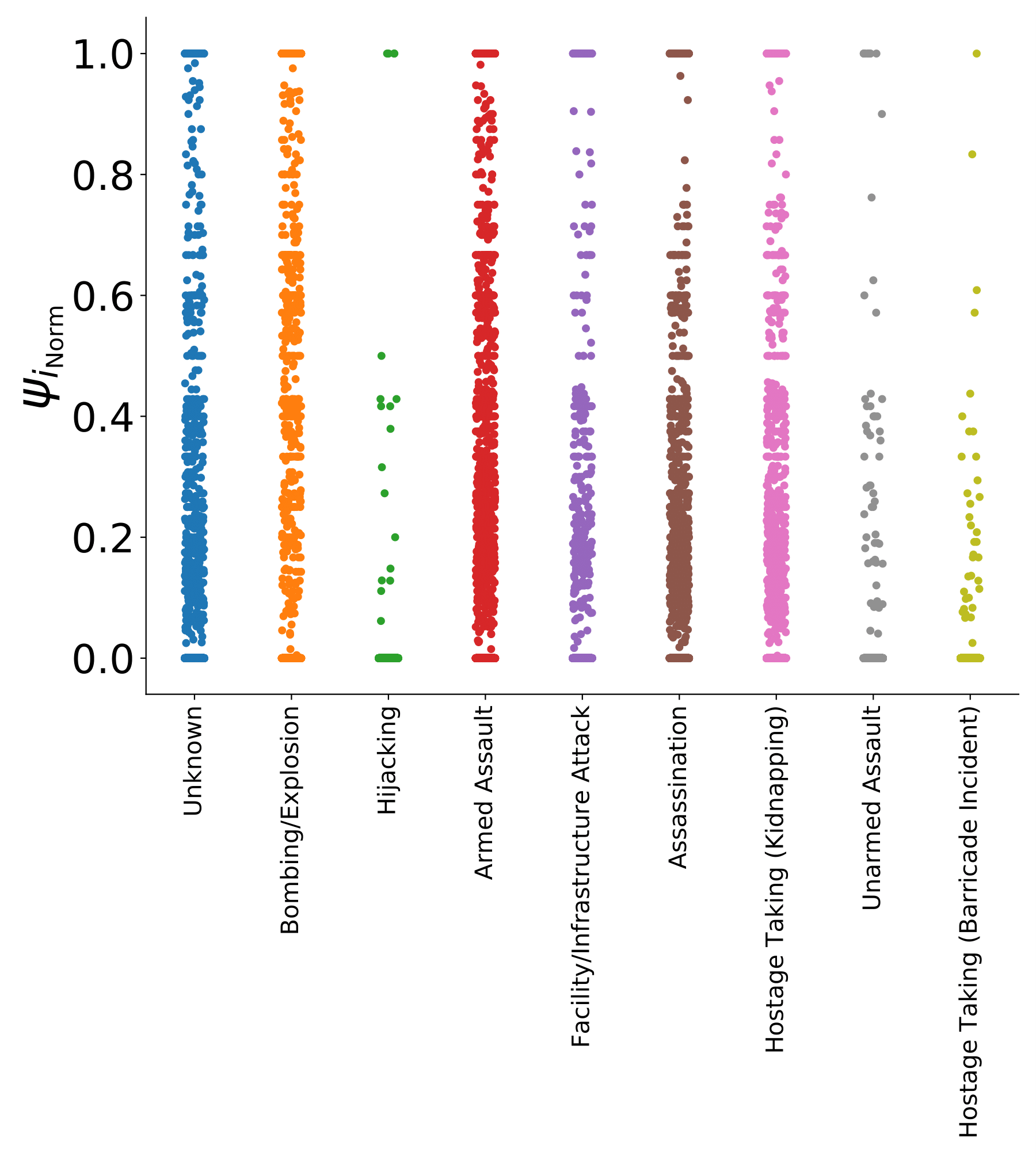}
    \caption{Distribution of $\psi_{i_{\mathrm{Norm}}}[u]$ over $U$ for all the tactics features $X$ - Afghanistan.}
    \label{fig:afg_tactics}
\end{figure}

All tactics are kept in the experimental analyses. Figure \ref{fig:afg_tactics} showcases the distribution of each tactic in terms of centrality values.

\subsubsection{Weapons}
A total of 7 weapon categories were utilized by terrorists in Afghanistan, these are: \textit{Unknown	Explosives,	Firearms, Incendiary,	Melee,	Sabotage Equipment,	Other.} Below we report the distribution of these weapons (Table \ref{afg_weap_tab}). 

\begin{table}[!hbt]
\centering
\footnotesize
\begin{tabular}{l c c}
\hline
\textbf{Weapon $x_j$} & \textbf{N of $\psi_{j_{\mathrm{Norm}}}[u]>0$} & \textbf{\%} \\ \hline
Explosives & 2020 & 61.42\% \\ \hline
Firearms & 1624 & 49.38\% \\ \hline
Unknown.1 & 876 & 26.63\% \\ \hline
Incendiary & 321 & 9.76\% \\ \hline
Melee & 194 & 5.90\% \\ \hline
Other & 6 & 0.18\% \\ \hline
Sabotage Equipment & 4 & 0.12\% \\ \hline
\end{tabular}
\caption{Afghanistan Weapons - Number of non-zero $\psi_{j_{\mathrm{Norm}}}[u]$  occurrences and percentage over $U$}
\label{afg_weap_tab}
\end{table}

\begin{figure}[!hbt]
    \centering
    \includegraphics[scale=0.10]{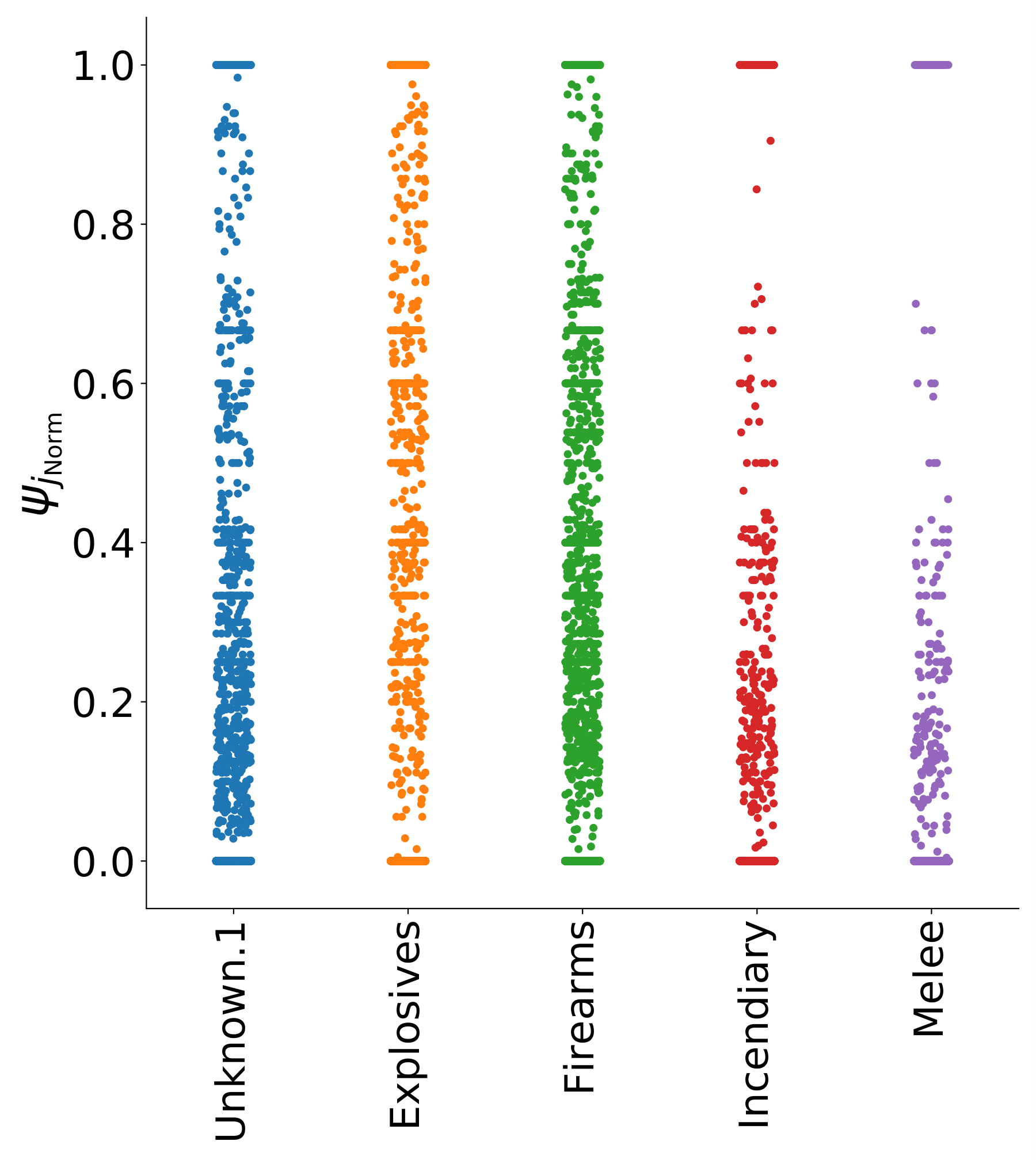}
    \caption{Distribution of $\psi_{j_{\mathrm{Norm}}}[u]$ over $U$ for all the weapons features $W$ - Afghanistan.}
    \label{fig:afg_wea}
\end{figure}

\textit{Other} and \textit{Sabotage Equipment} have been filtered out from the final set of weapons given their extremely low prevalence, leading to a total of 5 weapons. The centrality value distributions of the features in the final set of weapons $X$ are shown in Figure \ref{fig:afg_wea}.

\subsubsection{Targets}
In the 2000-2018 period, a total of 21 target categories were hit at least once by terrorists: \textit{Terrorists/Non-State Militia,	Government (General),	Private Citizens \& Property,	Airports \& Aircraft,	Unknown	Government, (Diplomatic) Journalists \& Media,	Police,	Business,	Religious Figures/Institutions,	Military,	Educational Institution,	Telecommunication,	Transportation,	NGO,	Food or Water Supply,	Tourists,	Utilities,	Other,	Violent Political Party,	Maritime}. Below Table \ref{afg_tar_tab} shows the distribution of these targets.

\begin{table}[!hbt]
\centering
\footnotesize
\begin{tabular}{l c c}
\hline
\textbf{Target $y_k$} & \textbf{N of $\psi_{k_{\mathrm{Norm}}}[u]>0$} & \textbf{\%} \\ \hline
Private Citizens \& Property & 1635 & 49.71\% \\ \hline
Police & 1506 & 45.79\% \\ \hline
Government (General) & 1151 & 35.00\% \\ \hline
Military & 518 & 15.75\% \\ \hline
Unknown.2 & 505 & 15.35\% \\ \hline
Business & 465 & 14.14\% \\ \hline
Educational Institution & 316 & 9.61\% \\ \hline
Religious Figures/Institutions & 231 & 7.02\% \\ \hline
NGO & 155 & 4.71\% \\ \hline
Terrorists/Non-State Militia & 154 & 4.68\% \\ \hline
Transportation & 144 & 4.38\% \\ \hline
Government (Diplomatic) & 133 & 4.04\% \\ \hline
Journalists \& Media & 82 & 2.49\% \\ \hline
Airports \& Aircraft & 54 & 1.64\% \\ \hline
Telecommunication & 53 & 1.61\% \\ \hline
Utilities & 38 & 1.16\% \\ \hline
Violent Political Party & 21 & 0.64\% \\ \hline
Food or Water Supply & 15 & 0.46\% \\ \hline
Tourists & 4 & 0.12\% \\ \hline
Maritime & 1 & 0.03\% \\ \hline
\end{tabular}
\caption{Afghanistan Targets - Number of non-zero $\psi_{k_{\mathrm{Norm}}}[u]$  occurrences and percentage over $U$}
\label{afg_tar_tab}
\end{table}

\begin{figure}[!hbt]
    \centering
    \includegraphics[scale=0.10]{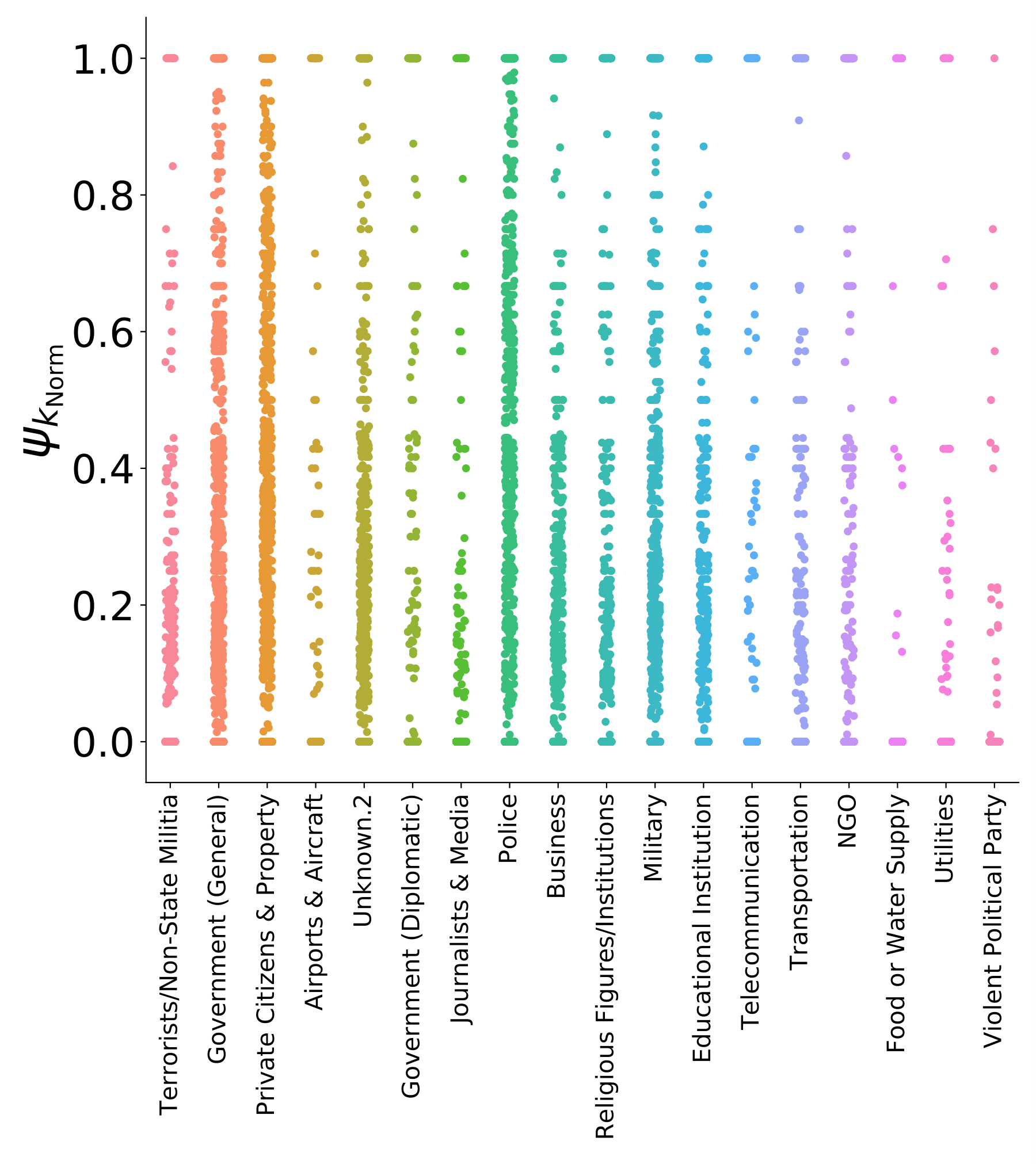}
    \caption{Distribution of $\psi_{k_{\mathrm{Norm}}}[u]$ over $U$ for all the target features $Y$ - Afghanistan.}
    \label{fig:afg_targets}
\end{figure}

Given the low prevalence, as done for all the other dimensions (in both datasets) we have proceeded to exclude those features that were present less than 10 times of the course of the entire 2000-2018 period. \textit{Tourists} and \textit{Maritime} are thus excluded from the experiments, leading to a total of 18 targets (Figure \ref{fig:afg_targets}).

\subsection{Iraq}
\begin{figure}
    \centering
    \includegraphics[scale=0.35]{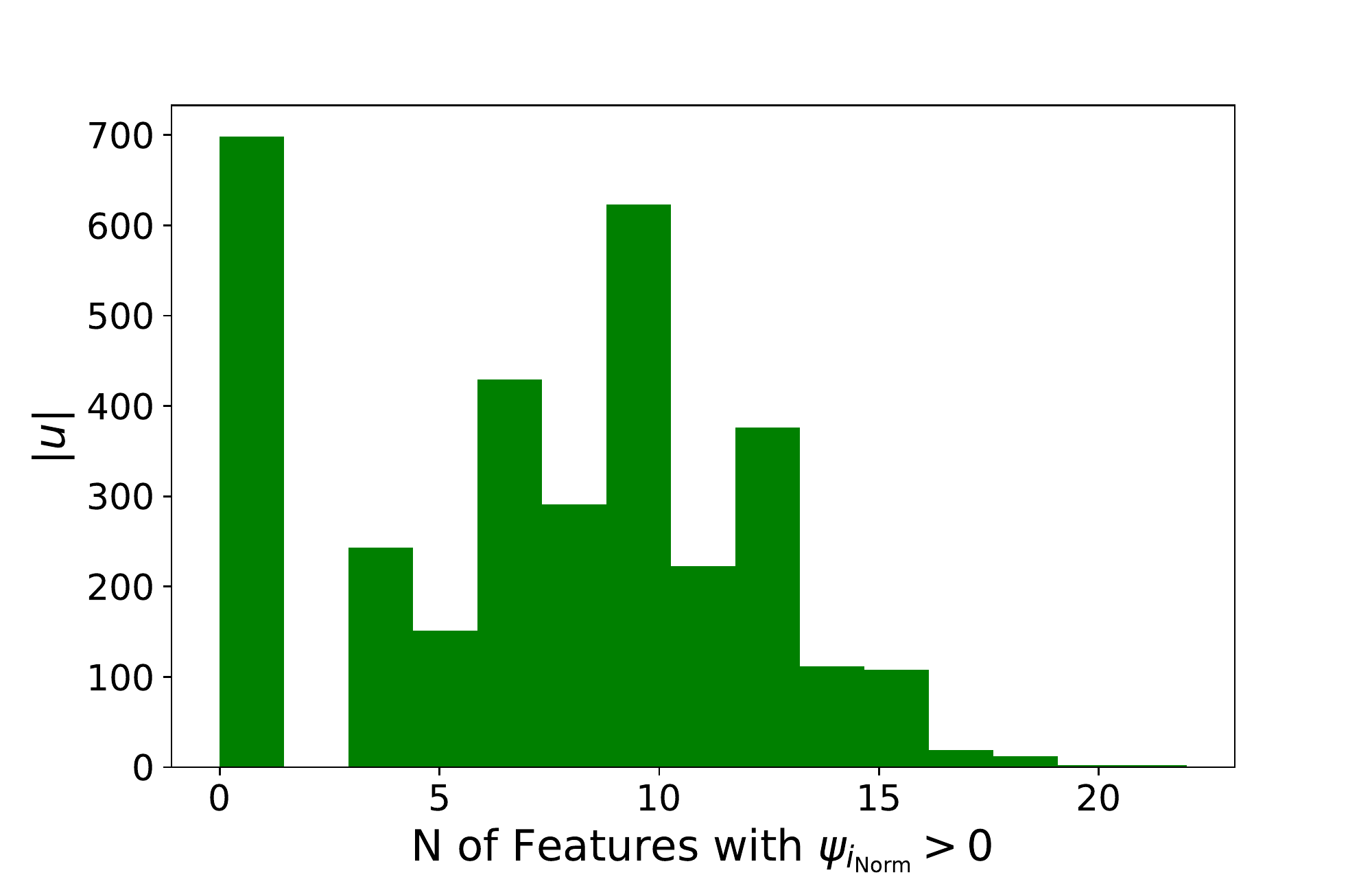}
    \caption{Count of non-zero $\psi_{i_{\mathrm{Norm}}}[u]$ in each $u$ - Iraq. The figure has been created using Matplotlib version 3.1.3.}
    \label{fig:nonzero_ira}
\end{figure}

In the Iraq case, out of the 3,289 time units, 698 recorded no attacks (21.2\%), thus having all the associated time series with 0 as inputs. Below, the histogram reporting the count of non-zero features for all the multivariate time-series at each time unit $u$ (Figure \ref{fig:nonzero_ira}). 
\newpage

\subsubsection{Tactics}
In Iraq, terrorists have employed 9 distinct tactics from 2000 to 2018: \textit{Assassination,	Bombing/Explosion,	Unknown	Armed Assault,	Facility/Infrastructure Attack,	Hostage Taking (Kidnapping),	Unarmed Assault,	Hostage Taking (Barricade Incident),	Hijacking} . The distribution is reported below (Table \ref{ira_tac_tab}).

\begin{table}[!hbt]
\centering
\footnotesize
\begin{tabular}{l c c}
\hline
\textbf{Tactic $x_i$} & \textbf{N of $\psi_{i_{\mathrm{Norm}}}[u]>0$} & \textbf{\%} \\ \hline
Bombing/Explosion & 2459 & 74.76\% \\ \hline
Armed Assault & 1440 & 43.78\% \\ \hline
Assassination & 863 & 26.24\% \\ \hline
Hostage Taking (Kidnapping) & 612 & 18.61\% \\ \hline
Unknown & 394 & 11.98\% \\ \hline
Facility/Infrastructure Attack & 136 & 4.13\% \\ \hline
Hostage Taking (Barricade Incident) & 25 & 0.76\% \\ \hline
Unarmed Assault & 10 & 0.30\% \\ \hline
Hijacking & 7 & 0.21\% \\ \hline
\end{tabular}
\caption{Iraq Tactics - Number of non-zero $\psi_{i_{\mathrm{Norm}}}[u]$  occurrences and percentage over $U$}
\label{ira_tac_tab}
\end{table}

\begin{figure}[!hbt]
    \centering
    \includegraphics[scale=0.10]{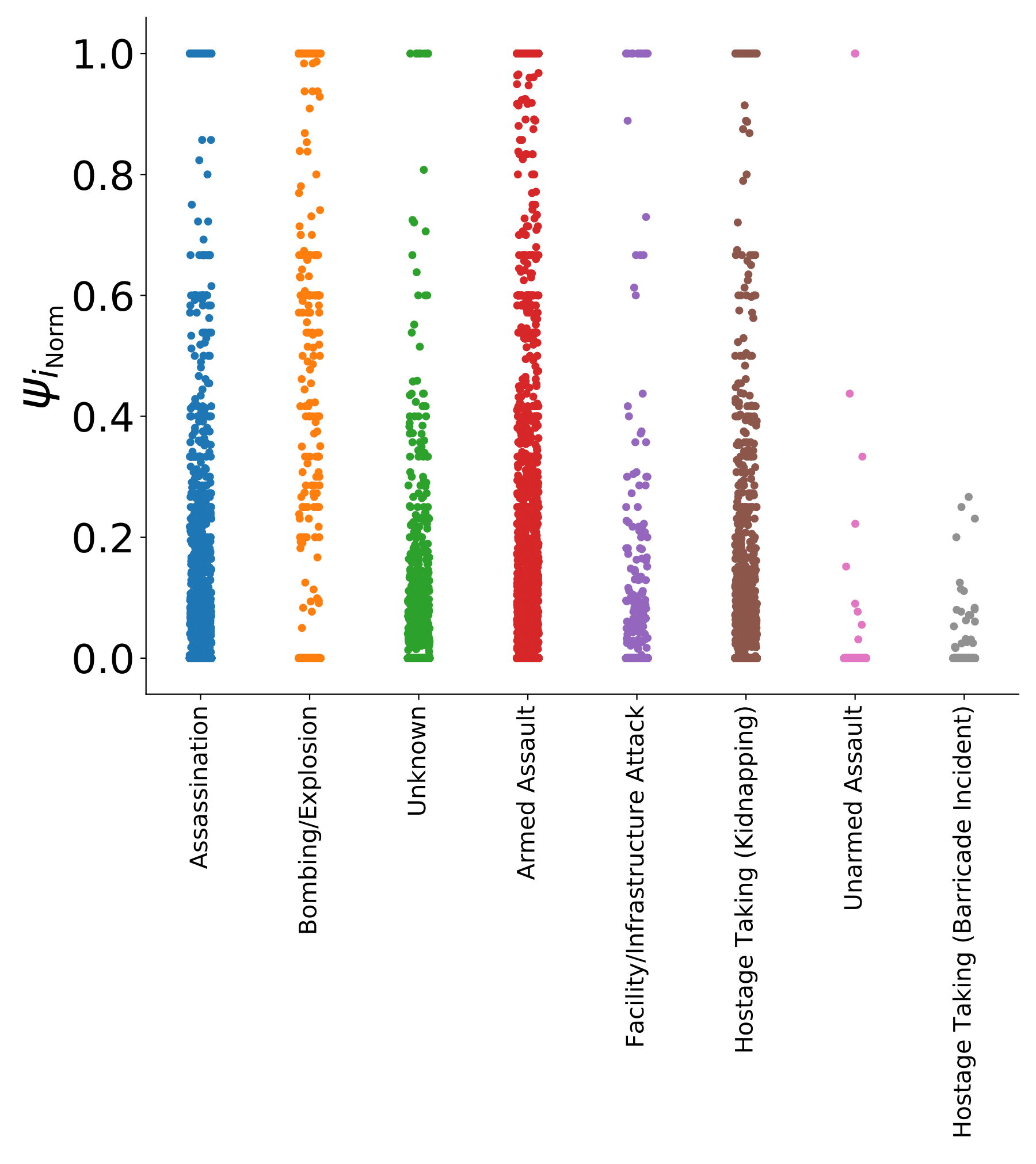}
    \caption{Distribution of $\psi_{i_{\mathrm{Norm}}}[u]$ over $U$ for all the tactics features $X$ - Iraq. The figure has been created using Matplotlib version 3.1.3.}
    \label{fig:ira_tactics}
\end{figure}

\textit{Hijacking} has been excluded from the multivariate time-series given the extremely low prevalence over the entire history $U$. The distribution of the values of features in $X$ for Iraq are displayed below (Figure \ref{fig:ira_tactics}).

\subsubsection{Weapons}
Overall, 7 weapon categories have at least one occurrence in the Iraq dataset: \textit{Explosives, Firearms, Unknown, Incendiary, Melee, Other, Sabotage Equipment} (Table \ref{ira_weap_tab}).

\begin{table}[!hbt]
\centering
\footnotesize
\begin{tabular}{l c c}
\hline
\textbf{Weapon $x_j$} & \textbf{N of $\psi_{j_{\mathrm{Norm}}}[u]>0$} & \textbf{\%} \\ \hline
Explosives & 2476 & 75.28\% \\ \hline
Firearms & 1739 & 52.87\% \\ \hline
Unknown.1 & 635 & 19.31\% \\ \hline
Incendiary & 107 & 3.25\% \\ \hline
Melee & 100 & 3.04\% \\ \hline
Other & 9 & 0.27\% \\ \hline
Sabotage Equipment & 5 & 0.15\% \\ \hline
\end{tabular}
\caption{Iraq Weapons - Number of non-zero $\psi_{j_{\mathrm{Norm}}}[u]$  occurrences and percentage over $U$}
\label{ira_weap_tab}
\end{table}

\begin{figure}[!hbt]
    \centering
    \includegraphics[scale=0.10]{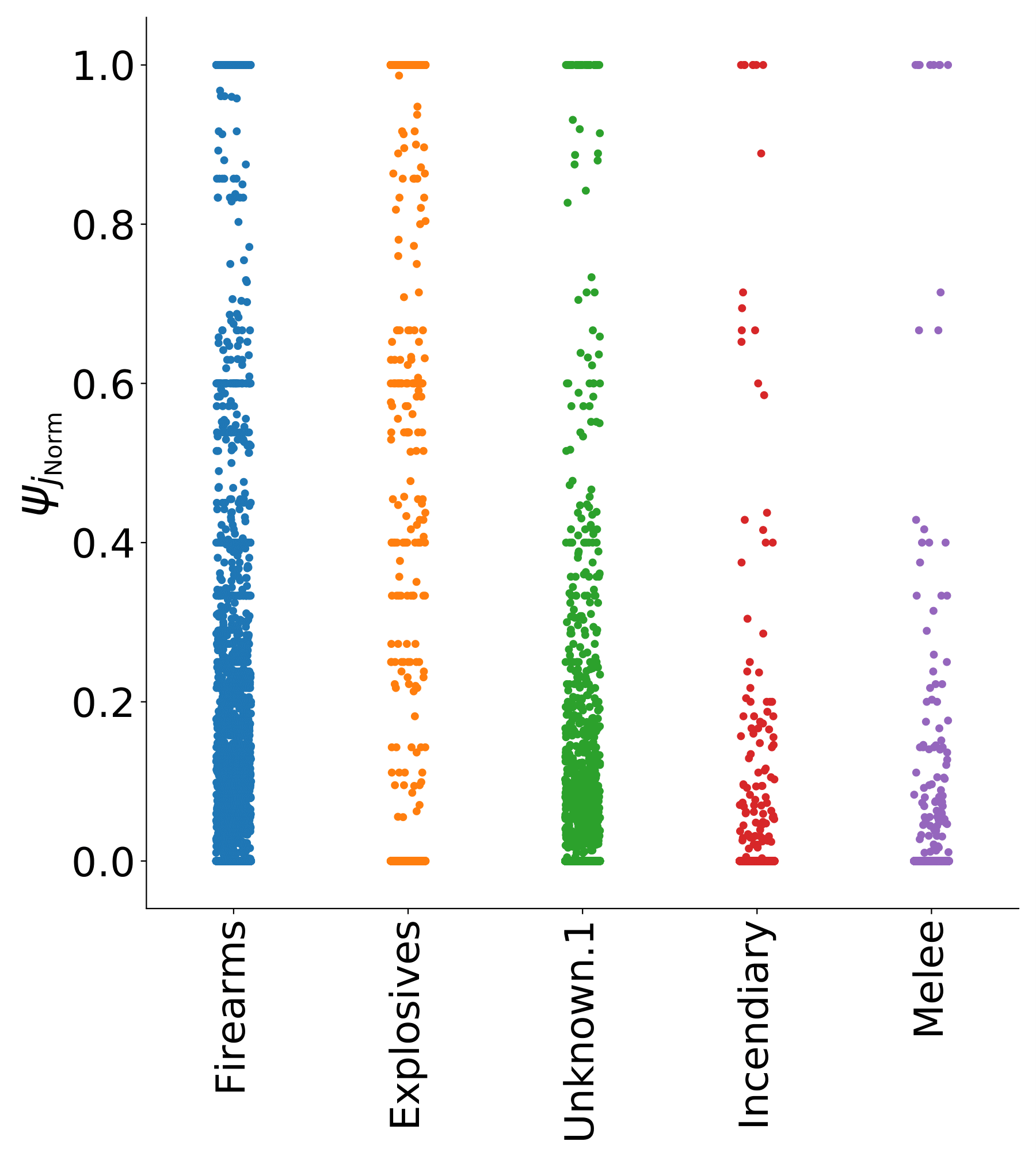}
    \caption{Distribution of $\psi_{j_{\mathrm{Norm}}}[u]$ over $U$ for all the weapon features $W$ - Iraq. The figure has been created using Matplotlib version 3.1.3.}
    \label{fig:ira_weapons}
\end{figure}

As done for the Afghanistan dataset, also in this case \textit{Other} and \textit{Sabotage Equipment} have been removed from the set of multivariate time series. The final set of $W$ for Iraq is displayed in Figure \ref{fig:ira_weapons}.

\subsubsection{Targets}
Overall, in Iraq 21 target types have been hit from 2000 to 2018. These are: \textit{Private Citizens \& Property,	Government (Diplomatic),	Business,	Police,	Government (General),	NGO,	Journalists \& Media,	Violent Political Party,	Religious Figures/Institutions,	Transportation,	Unknown,	Terrorists/Non-State Militia,	Utilities,	Military,	Telecommunication,	Educational Institution,	Maritime,	Tourists,	Other,	Food or Water Supply,	Airports \& Aircraft}. The distribution of occurrences is displayed below (Table \ref{ira_targ_tab}). 

\begin{table}[!hbt]
\centering
\footnotesize
\begin{tabular}{l c c}
\hline
\textbf{Target $y_k$} & \textbf{N of $\psi_{k_{\mathrm{Norm}}}[u]>0$} & \textbf{\%} \\ \hline
Private Citizens   \& Property & 2220 & 67.50\% \\ \hline
Police & 1602 & 48.71\% \\ \hline
Government (General) & 1326 & 40.32\% \\ \hline
Business & 1210 & 36.79\% \\ \hline
Terrorists/Non-State Militia & 747 & 22.71\% \\ \hline
Military & 599 & 18.21\% \\ \hline
Unknown.2 & 566 & 17.21\% \\ \hline
Religious Figures/Institutions & 487 & 14.81\% \\ \hline
Transportation & 425 & 12.92\% \\ \hline
Educational Institution & 250 & 7.60\% \\ \hline
Utilities & 233 & 7.08\% \\ \hline
Journalists \& Media & 168 & 5.11\% \\ \hline
Government (Diplomatic) & 105 & 3.19\% \\ \hline
Violent Political Party & 65 & 1.98\% \\ \hline
Other.1 & 57 & 1.73\% \\ \hline
Food or Water Supply & 28 & 0.85\% \\ \hline
Airports \& Aircraft & 27 & 0.82\% \\ \hline
NGO & 25 & 0.76\% \\ \hline
Telecommunication & 19 & 0.58\% \\ \hline
Tourists & 10 & 0.30\% \\ \hline
Maritime & 2 & 0.06\% \\ \hline
\end{tabular}
\caption{Iraq Targets - Number of non-zero $\psi_{k_{\mathrm{Norm}}}[u]$  occurrences and percentage over $U$}
\label{ira_targ_tab}
\end{table}

\begin{figure}[!hbt]
    \centering
    \includegraphics[scale=0.10]{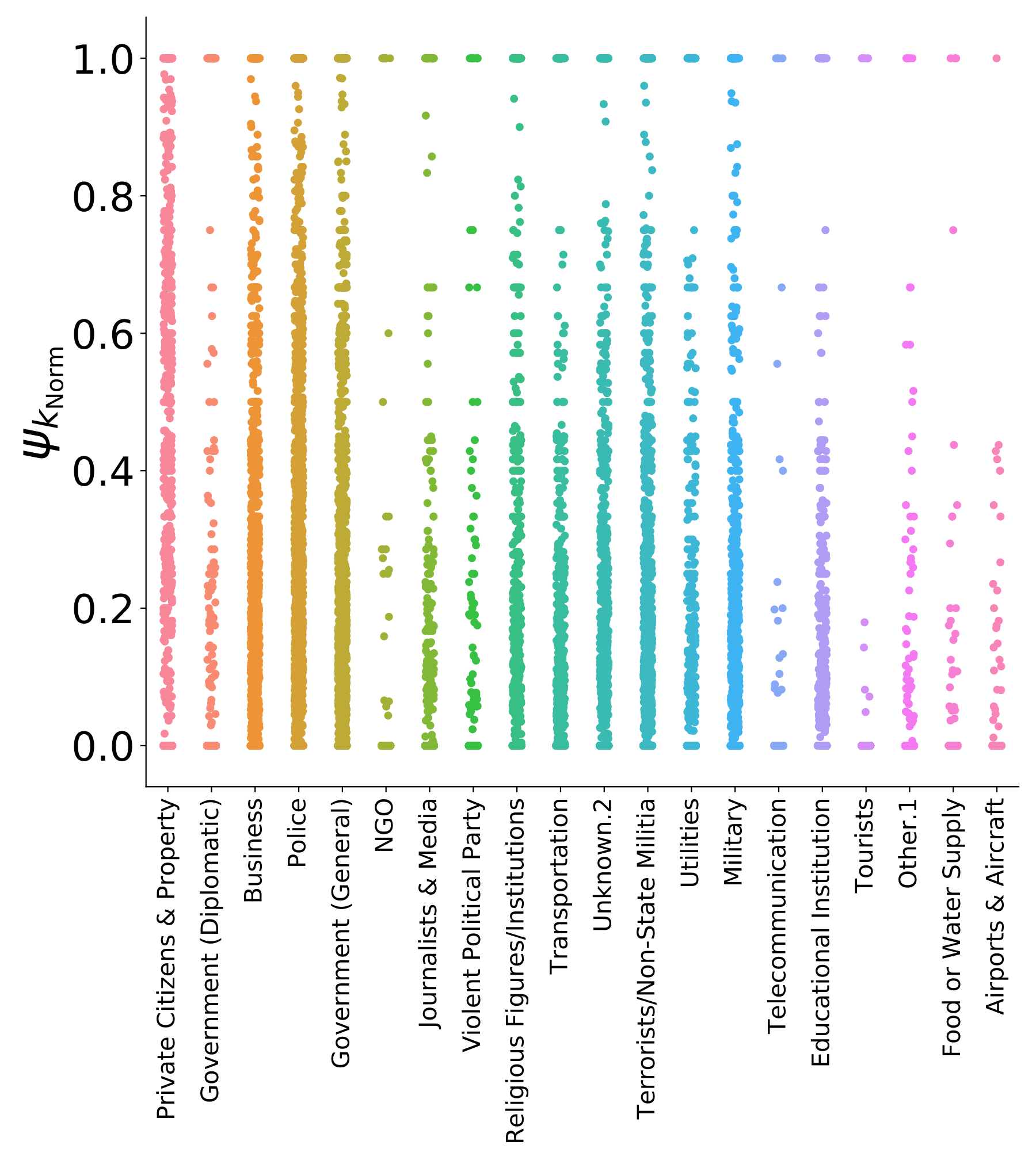}
    \caption{Distribution of $\psi_{k_{\mathrm{Norm}}}[u]$ over $U$ for all the target features $Y$ - Iraq. The figure has been created using Matplotlib version 3.1.3.}
    \label{fig:ira_targets}
\end{figure}

Given the low prevalence, \textit{Maritime} has been excluded, leading to a total of 20 time series mapping targets in Iraq (Figure \ref{fig:ira_targets}).

\section{Experiments}

\subsection{Algorithms: Architectures' Details}
This subsection provides details on the architectures of the different models, except the baseline which did not involve any learning mechanism. Overall, all models have been trained for 100 epochs using Adam as the optimizer given its ability in noisy problems involving sparse gradients \citep{KingmaAdamMethodStochastic2017} and a batch size equal to 16. Furthermore, we set 10 as the patience hyper-parameter mapping the validation loss of the model. As described in the main manuscript, all models have been run using different input width in terms of time units $u$, to understand what  the optimal length of the recent history to take into account in order to obtain better forecasts is:\footnote{The only exception is the Baseline model that, given its particular architecture, only used the previous time unit to infer forecasts at $u+1$.} these input widths were 1 (=2 days), 5 (=10 days), 15 (=1 month), 30 (=2 months).

It will be noted that the architectures are not particularly complex, i.e., they do not involve multiple hidden layers in most cases: this is mostly due to the limited amount of data (in terms of $u$) at our disposal. We also performed experiments with more complex architectures made of a higher numbers of stacked layers (and higher number of units and filters), but the complexity of the networks did not lead to increments in algorithmic performance (while leading instead in higher computational costs).  Nonetheless, the resulting outcomes of the models presented in the paper indicate that even simple learning architectures are capable of efficiently forecasting terrorist targets. This aspect inserts in an emerging area of research that investigates the benefits of training simpler models over massive networks \citep{BaDeepNetsReally2014}. Instead of being a limitation, reaching good performance with simple and computationally cheap models may be considered a strength of the proposed computational framework and, particularly, of the engineering of our feature space.  As a final note, in addition to the specific use of Dropout as a regularized for some models, all architectures have been trained using early stopping with a patience of 10 epochs in relation to Mean Squared Error to further limit the risk of overfitting.  

\subsubsection{Feedforward Neural Network (FNN)}
The FNNs trained in our experiments involved an input layer, followed by a flatten layer with 0 trainable parameters. Following, two dense layers with 32 neurons each with Rectified Linear Unit (ReLU) as the activation function. Finally, the last layer involved a number of units equal to the number of targets (i.e., 18 in the Afghanistan dataset, 20 in the Iraq one), and was followed by a reshape layer with no parameters. All dense layers had Glorot uniform as the kernel initializer and included a bias vector. 
\subsubsection{Long Short-Term Memory Network (LSTM)}
The LSTM networks involved a first input layer, followed by an LSTM layer with 32 neurons and a 0.5 dropout to avoid overfitting \citep{SrivastavaDropoutSimpleWay2014a}. The LSTM layer used Tanh as the activation function (given the need to forecast values in the $[0,1]$ range) and Glorot uniform as the kernel initializer. The recurrent activation function was a sigmoid and the LSTM layer also included a bias vector. The recurrent initializer was an orthogonal matrix to prevent vanishing and exploding gradients \citep{LeSimpleWayInitialize2015,HenaffRecurrentOrthogonalNetworks2017}.  Finally, the last layer was a dense one having as units the number of targets in each dataset, and the same hyperparameters as the dense layers described in the FNN network.

\subsubsection{Convolutional Neural Network (CNN)}
The trained CNNs have a total of four layers. Besides the first input one, we have included a 1D-Convolutional layer with 32 filters, followed by two dense layers (one with 32 units, the other with a number of units equal to $|Y|$), sharing the same hyperparameters of the dense layers in the previous models. The 1D-Convolutional layer made use of a bias vector with no padding. The activation function used in the 1D-Convolutional layer and the first dense layers was a ReLU. 
\subsubsection{Bi-Directional Long Short-Term Memory Network (Bi-LSTM)}
The Bi-LSTM networks are very similar in their architecture to the LSTM ones. The only (relevant) difference is that instead of having a simple LSTM layer, it actually has a bidirectional one (involving a double number of parameters), with concatenation as the merge mode. The dropout is set as 0.5. The Bidirectional layer uses Tanh as activation and sigmoid as the recurrent activation function and allows the network to access the hidden state output at each input time unit.
\subsubsection{CNN-LSTM (CLDNN)}
Finally, the CNN-LSTM model that takes inspiration from the CLDNN architecture proposed by \citeNew{SainathConvolutionalLongShortTerm2015}, was engineered with a first input layer, followed by a 1D-Convolutional layer with 32 filters, a 1D-Max Pooling layer with pool size equal to 2, a dense layer with 32 units with ReLU as activation, an LSTM layer with 32 neurons, 0.5 dropout and Tanh as activation and a final dense layer with a number of units equal to $Y$. In the first layer (1D-Convolutional) there is no padding and Glorot uniform is used as the kernel initializer. In the 1D-Max Pooling layer also no padding is performed. The dense and LSTM remaining layers share the same hyperparameters of the previously outline dense and LSTM layers found in the other models.

\section{Additional Results}

\begin{figure}[!hbt]
    \centering
    \includegraphics[scale=0.4]{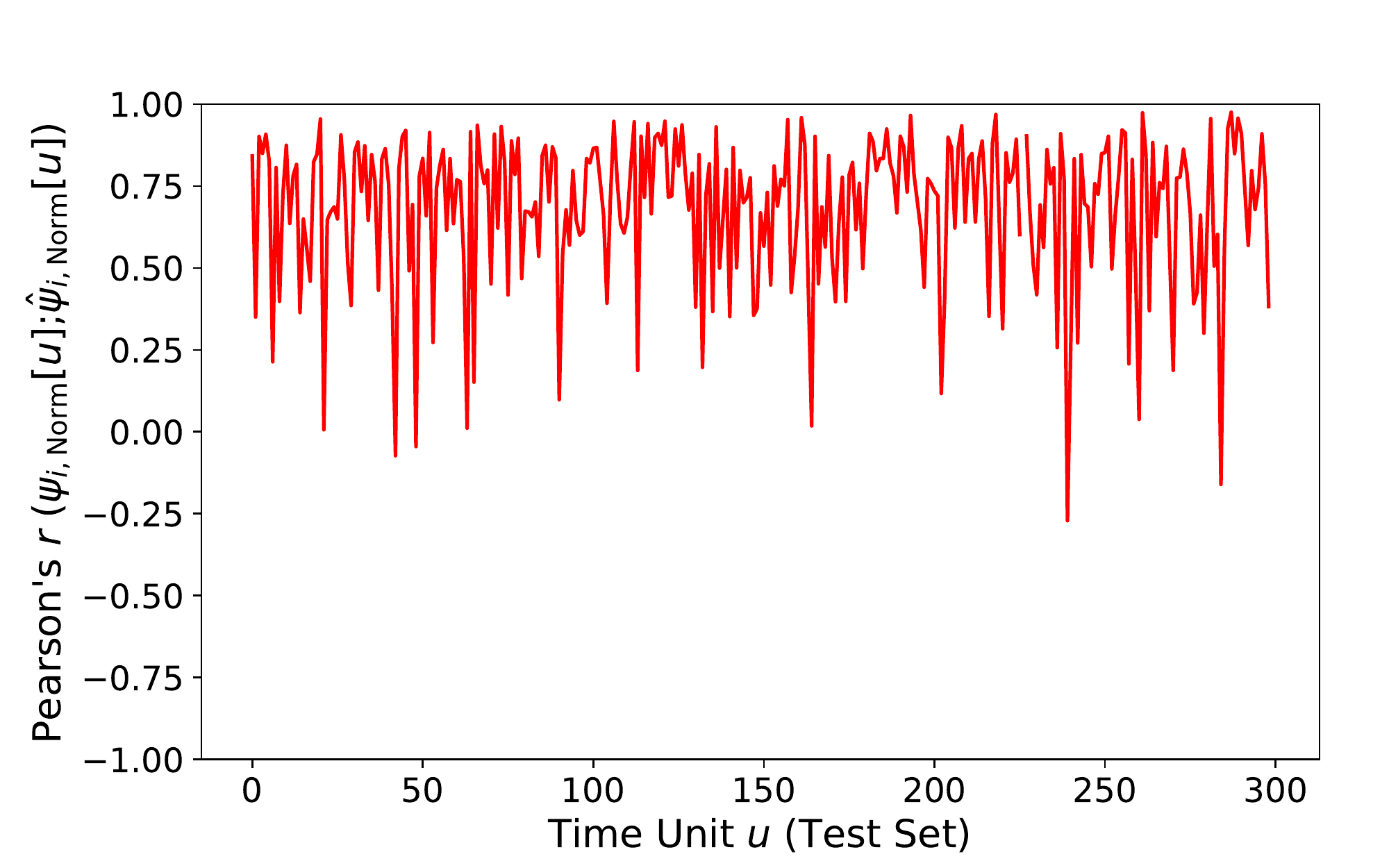}
    \caption{Correlation between empirical and forecasted centrality (test set) - Afghanistan.}
    \label{fig:cor_afg}
\end{figure}

\begin{figure}[!hbt]
    \centering
    \includegraphics[scale=0.4]{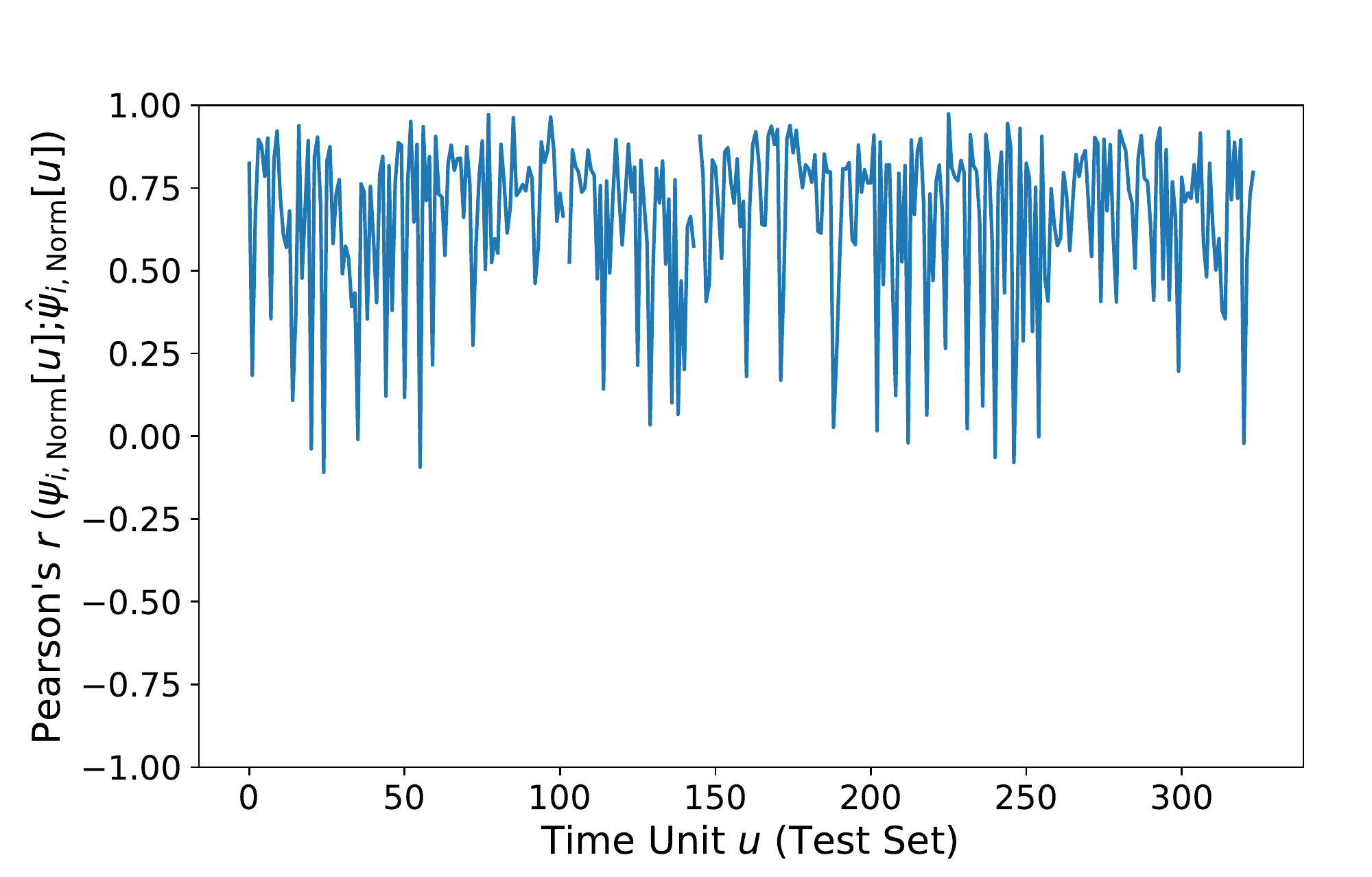}
    \caption{Correlation between empirical and forecasted centrality (test set) - Iraq.}
    \label{fig:cor_iraq}
\end{figure}

Figures \ref{fig:cor_afg} and \ref{fig:cor_iraq} integrate the results presented in the main text, highlighting the correlation between each vector $\Psi_{norm}[u]$ and $\hat{\Psi}_{norm}[u]$, representing respectively the empirical centrality values of each target feature at each time stamp $u$ and the corresponding predicted centrality values. Both graphs relates to the models reaching the highest $\Gamma$ for the Afghanistan and Iraq cases.

While both correlation signals suffer from oscillations that, in certain cases, lead to unsatisfactory correlation values that get close to zero or become even negative, the mean correlation value for Afghanistan is 0.71 (with SD=0.23), while for Iraq is 0.69 (with SD=0.21). The first moment of both distributions corroborates the ability of the models with the highest performance in capturing the underlying dynamics and trend found in the empirical multivariate time-series. 

Nonetheless, future work should investigate the reason of the negative oscillations to understand what causes them, while concurrently improve the overall forecasting performance at the general level (in terms of $\Gamma$ and $\Phi$) and at the target-level, to guarantee model interpretability and results validity, two core challenges in applied deep learning in many fields. 

\newpage

\bibliography{AAAI21}

\bibliographyNew{AAAI21_appendix}

\end{document}